\documentclass{article}

    \PassOptionsToPackage{numbers, compress}{natbib}

\usepackage[final]{neurips_2025}

\usepackage{amsfonts}       
\usepackage{amsmath}
\usepackage{amssymb}
\usepackage{amsthm}
\usepackage{array}
\usepackage{algorithm}
\usepackage{algpseudocode}
\usepackage{booktabs}       
\usepackage{bm}
\usepackage{colortbl}
\usepackage{caption}
\usepackage{float}
\usepackage{hyperref}       
\usepackage[capitalize,noabbrev]{cleveref}
\usepackage[utf8]{inputenc}
\usepackage[T1]{fontenc}    
\usepackage{graphics}       
\usepackage[utf8]{inputenc} 
\usepackage{microtype}      
\usepackage{mathtools}
\usepackage{multirow}
\usepackage{nicefrac}       
\usepackage{subfigure}
\usepackage{subcaption} 
\usepackage{tabularx}
\usepackage[textsize=tiny]{todonotes}
\usepackage{url}            
\usepackage{wrapfig}

\title{RSAVQ: Riemannian Sensitivity-Aware Vector Quantization for Large Language Models }

\author{%
  Zukang Xu\thanks{Equal contribution.}\\
  Houmo AI\\
  \And
  Xing Hu\footnotemark[1]\\
  Houmo AI\\
  \And
  Qiang Wu\\
  Houmo AI\\
  \And
  Dawei Yang\thanks{Corresponding author: \texttt{dawei.yang@houmo.ai}}\\
  Houmo AI\\
}

\begin{document}

\maketitle
\begin{abstract}
Large language models (LLMs) have demonstrated remarkable performance across a wide range of natural language processing tasks. However, their exponentially increasing parameters pose significant challenges for deployment on resource-constrained devices. 
Vector Quantization (VQ) shows great promise for low-bit quantization (e.g., 2 to 4 bits), but existing work faces two key challenges: unconstrained direction error and suboptimal bit allocation.
In this paper, we propose RSAVQ, a novel VQ framework to enhance extremely low-bit quantization for LLMs. RSAVQ introduces two geometry-driven innovations that effectively mitigate above limitations:
(1) Error Direction Sensitivity Guidance (EDSG), which leverages the Fisher Information Matrix (FIM)-induced Riemannian metric to project quantization errors onto low-sensitivity directions in the parameter space. Specifically, this projection is performed along the negative natural gradient direction, which effectively suppresses error expansion.
(2) Weight Channel Sensitivity Guidance (WCSG) , which constructs a channel-wise sensitivity metric via FIM curvature analysis to dynamically guide bit resource allocation. The approach facilitates a globally optimal quantization solution within prescribed bit constraints. 
Experiments demonstrate that RSAVQ outperforms existing methods for LLMs. For example, in 2-bit quantization of LLaMA-3 8B, RSAVQ leads baselines like VPTQ and QuIP\# by 0.4 in perplexity (PPL) and 1.5 in zero-shot accuracy. 
This work offers a practical solution for constrained environments and a theoretical bridge between information geometry and the quantization of neural networks, advancing efficient deep learning.
\end{abstract}
\section{Introduction}\label{sec_intro}

In recent years, large language models (LLMs) have achieved breakthrough results in natural language processing (NLP), code generation, reasoning, and multimodal tasks\citep{llama2,llama3,llama4,llava,qwen}.
While this progress is impressive, it comes at the cost of a dramatic increase in model size (e.g., the LLaMA-3 70B\citep{llama3} model demands around 140GB of memory in FP16 precision), posing significant barriers when deploying on resource-constrained devices. 

Post-Training Quantization (PTQ)\citep{ptq} has emerged as a promising technique for reducing the resource footprint of LLMs by converting model weights into lower-bit fixed-point representations without the need for retraining. A typical strategy in PTQ is Scalar Quantization (SQ), where each individual weight is quantized independently to a lower-bit value.
Recent work \citep{gptq, awq, illm, rwkvquant, moequant} has achieved near-original model accuracy with 4 bit quantization. However, due to the limitations of numerical representation, SQ struggles to maintain performance at extremely low bits (e.g. 3 bit or fewer), often leading to significant degradation in model accuracy.

\begin{figure}[htbp]
    \vspace{-0.45cm}
    \begin{minipage}[t]{0.4\textwidth}
        \centering
        \includegraphics[width=\linewidth]{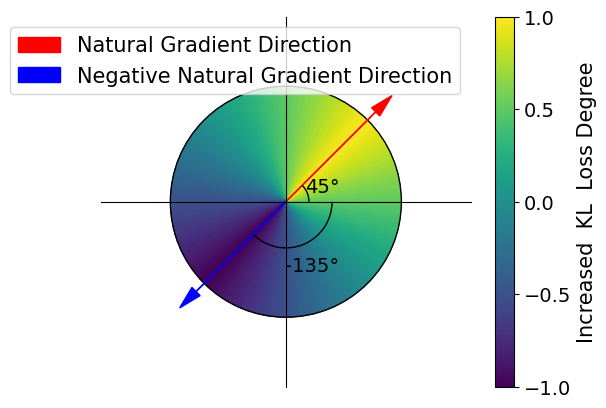}
        \caption{Schematic diagram of loss perturbation from isotropic weight perturbations: natural gradient direction yields maximum loss increment.}
        \label{fig1:Error direction sensitivity}
    \end{minipage}
    \hfill
    \begin{minipage}[t]{0.55\textwidth}
        \centering
        \includegraphics[width=\linewidth]{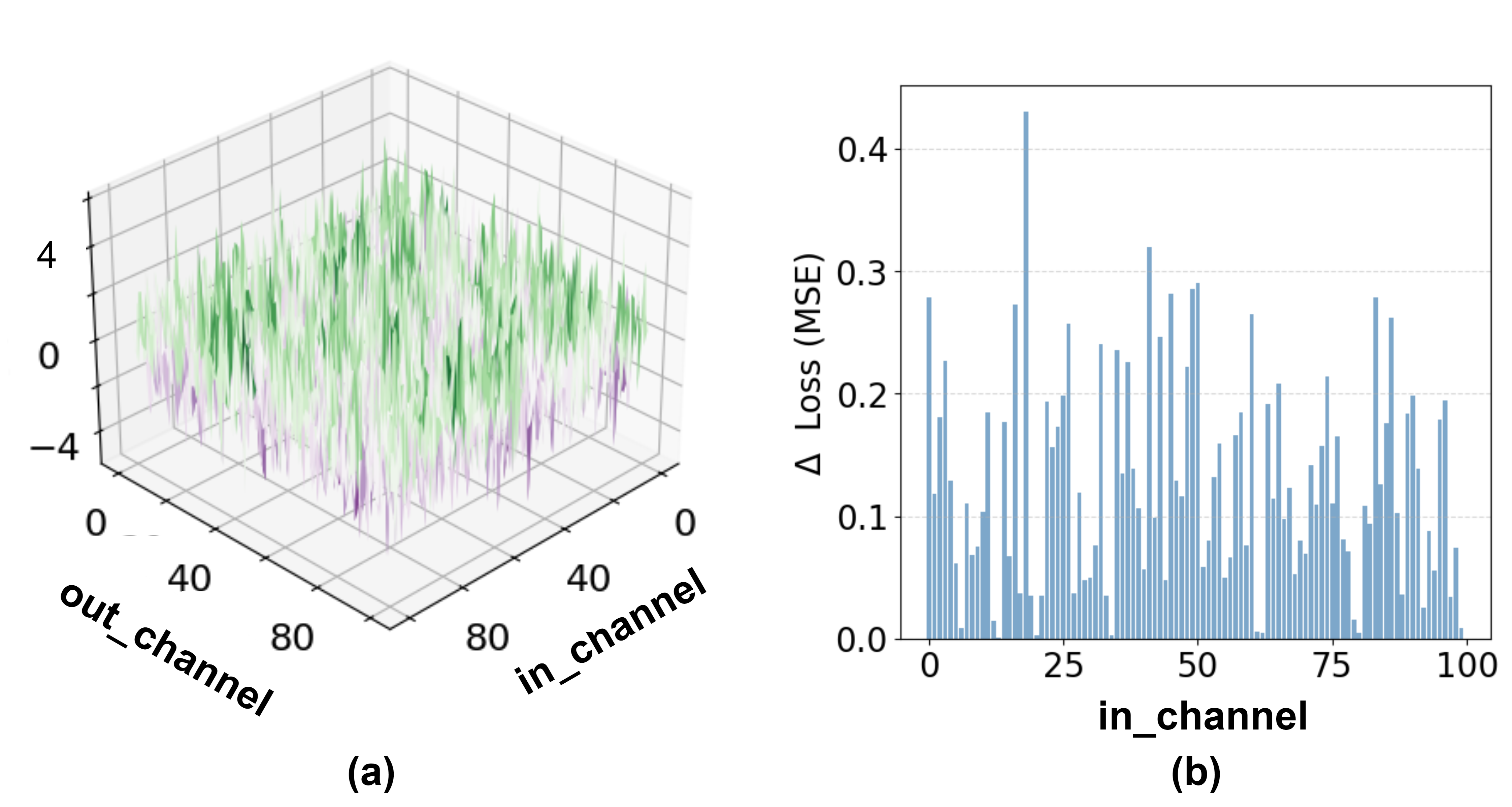}
        \caption{(a) Weight distribution of sampling patch in down-project layer of final block in LLaMA-3 8B model. (b) Change in loss after applying the same perturbation( + 0.05 per element) to the input channel of figure (a).}
        \label{fig2:channel-wise sensitivity}
    \end{minipage}
    \vspace{-0.5cm} 
\end{figure}

In contrast, Vector Quantization (VQ)\citep{vector_quantize}, another strategy in PTQ shows potential in ultra-low-bit LLM quantization. VQ maps high-dimensional vectors to a set of predefined lower-dimensional vectors and achieves more effective data compression than SQ by leveraging correlations and redundancies across different data dimensions. However, we identify that there are two critical limitations in existing VQ-based methods~\citep{quip_sharp, gptvq, vptq, aqlm, pcdvq}:
(1) \textbf{unconstrained direction error}: these methods often overlook directional discrepancies between floating-point vectors and their quantized representations, which significantly affect model performance.
In particular, by treating quantization errors as isotropic perturbations under Euclidean assumptions, they fail to capture the geometric relationship between the loss function's sensitivity and the directions of perturbation.
As demonstrated in Fig.~\ref{fig1:Error direction sensitivity}, errors along the negative natural gradient direction (\(\theta = -135^\circ\)) lead to loss reduction or significantly less loss increase compared to those along the natural gradient direction (\(\theta = 45^\circ\)). This highlights that error direction control is critical for preserving accuracy. 
(2) \textbf{suboptimal bit allocation}: these methods often assume uniform sensitivity across weight channels and assign equal bit-widths to each channel. However, as shown in Fig.~\ref{fig2:channel-wise sensitivity}(b), applying identical perturbations on each channel results in varying effects on model accuracy, indicating that such a naive bit-allocation policy compromises quantization performance.

\begin{figure}[!h]
    \centering
    \vspace{-0.3cm}
    \includegraphics[width=0.7\linewidth]{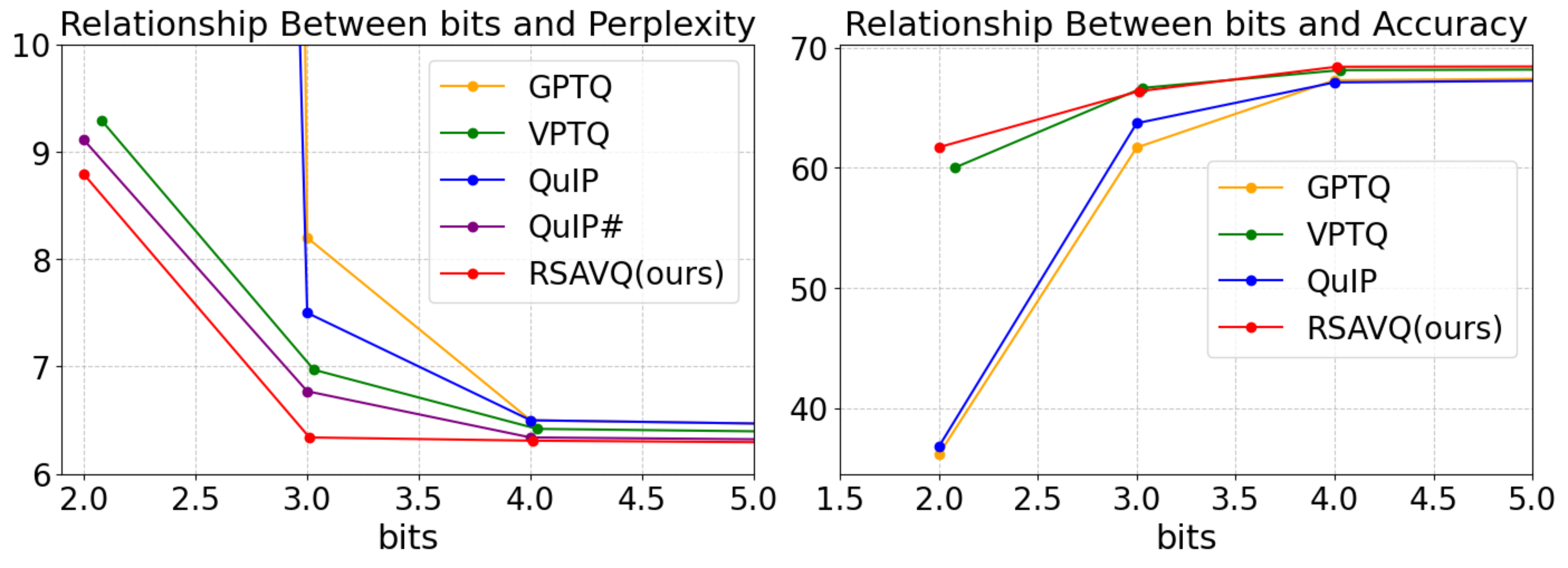}
    \caption{WikiText-2 PPL (left) and average zero-shot accuracy (right) for LLaMA-3 8B quantized at different bit-widths.}
    \vspace{-0.3cm}
    \label{fig3:method_compare}
\end{figure}

To address these limitations, we propose Riemannian Sensitivity-Aware Vector Quantization (RSAVQ), a novel VQ framework that leverages information geometry to model the parameter space (i.e., the weights) of LLMs  as a Riemannian manifold with non-uniform curvature—where the local geometry is described by Fisher Information Matrix (FIM) \citep{natural_gradient,riemannian}. RSAVQ employs FIM to characterize the local geometric structure of the parameter space, including inter-parameter correlations and manifold curvature, thereby enabling a precise quantification of how parameter perturbations along different directions affect the loss function.
RSAVQ comprises two core components:
(1)\textbf{Error Direction Sensitivity Guidance(EDSG)} projects inevitable quantization errors onto low-sensitivity directions (i.e., along the negative natural gradient directions on the Riemannian manifold), minimizing their adverse impact on model performance and effectively mitigating error accumulation.
(2)\textbf{Weight Channel Sensitivity Guidance (WCSG)} utilizes FIM to quantify each channel’s sensitivity, identifying functionally critical channels and dynamically allocating bit resources to prioritize high-curvature (sensitive) channels while applying aggressive compression to low-sensitivity counterparts.
This synergistic design enables RSAVQ to optimize the error direction of each quantized sub-vector through low-sensitivity projection and adaptively allocate bits based on channel-wise geometric analysis, thus achieving both ultra-low-bit compression and robust accuracy preservation in large language models.

Experiments demonstrate that RSAVQ achieves state-of-the-art performance on LLMs (see Fig.~\ref{fig3:method_compare} for detailed comparisons).
Specifically, for the LLaMA-3 8B model, our method significantly outperforms existing approaches across diverse quantization bit-widths. In particular, under 2-bit quantization, RSAVQ outperforms baselines such as VPTQ and QuIP\# by 0.4 in PPL and 1.5 in zero-shot accuracy.
Our main contributions are summarized as follows: 
\begin{itemize}
    \item We identify and systematically analyze two critical limitations in vector quantization—unconstrained direction error and suboptimal bit allocation—and demonstrate that they are key contributors of accuracy degradation in ultra-low-bit settings.
    \item We propose RSAVQ, a novel VQ framework grounded in information geometry. RSAVQ utilizes FIM to guide both error direction projection and per-channel bit allocation, integrating these insights into a single holistic algorithm.
    \item We validate RSAVQ on LLMs of various model sizes, demonstrating superior performance compared to existing quantization methods, particularly in extremely low-bit scenarios.
\end{itemize}

\vspace{-0.25cm}
\section{Related Work}\label{sec_related_work}
\vspace{-0.25cm}
Recent PTQ methods can be broadly categorized into scalar and vector quantization. These approaches typically assume a Euclidean parameter space, neglecting the intrinsic geometric structure of deep networks. 
Although geometric deep learning advances show that the parameter space is better modeled as a non-uniformly curved Riemannian manifold, these insights have only been theoretically explored, with their practical value in extreme low-bit quantization remaining unexplored. 
Moreover, to date, no unified framework has integrated information geometry to jointly optimize bit allocation and project errors onto low-sensitivity directions. This integration is crucial for minimizing performance degradation. Accordingly, we review related work from two angles: PTQ methods and geometric deep learning research. 

\vspace{-0.25cm}
\subsection{PTQ Methods}
\vspace{-0.25cm}
\paragraph{Scalar Quantization (SQ),} a classical quantization approach, maps weights to low-bit representations using fixed scaling factors and zero points. It relies on two key assumptions: (1) isotropic errors, where quantization errors in all directions equally impact model performance; (2) uniform parameter sensitivity, treating all parameters as requiring the same bit precision. 
Methods like GPTQ\citep{gptq} and AWQ\citep{awq} follow this paradigm, with GPTQ incorporating error compensation and AWQ adjusting outlier weights to stabilize quantization. Techniques such as Quarot\citep{quarot}, OSTQuant\citep{ostquant} and MambaQuant\citep{mambaquant} utilize Hadamard rotations to enhance weight distribution uniformity, yet these still operate under Euclidean assumptions. Critically, these SQ methods fail to model geometric sensitivities, leading to severe accuracy degradation in extreme low-bit scenarios (e.g. $\leq$ 3-bit) where direction-specific and channel-wise sensitivities dominate error impacts.
\vspace{-0.4cm}
\paragraph{Vector Quantization (VQ)}\citep{vector_quantize}methods enhance weight compression by clustering weight vectors into shared codebooks. For example, GPTVQ\citep{gptvq} integrates error compensation with EM algorithms to optimize codebooks and indices, while VPTQ\citep{vptq} and AQLM\citep{aqlm} adopt residual quantization to refine error fitting. CRVQ\citep{crvq} achieves 1-bit quantization by iteratively selecting critical channels for residual processing, and QuIP\# \citep{quip_sharp} uses Hadamard rotations to preprocess weights before uniform codebook quantization. Despite these advancements, existing VQ techniques lack adaptive bit allocation guided by channel sensitivity and fail to constrain quantization errors to low-impact geometric directions, leading to suboptimal performance in extreme low-bit scenarios.

\vspace{-0.25cm}
\subsection{Geometric Deep Learning}
\vspace{-0.25cm}
\paragraph{Geometric Deep Learning} highlights the non-Euclidean nature of neural network parameter spaces, suggesting that they can be more accurately modeled as Riemannian manifolds endowed with FIM\citep{geometric, natural_gradient, riemannian}.
Fisher Information Matrix (FIM)-based approaches\citep{fim} define a local Riemannian metric for the parameter space and demonstrate that natural gradient descent achieves faster convergence by adapting to the underlying manifold structure.
To address the computational complexity of FIM, methods such as K-FAC\citep{k-fac} propose efficient Kronecker-factored curvature approximations, making second-order optimization feasible for large-scale networks.
Beyond optimization, manifold-aware methods\citep{geometric_survey} have shown that leveraging geometric structures can further unlock hidden information in deep models through Riemannian updates.
\vspace{-0.25cm}
\paragraph{Explorations of Riemannian manifolds in quantization.}
\vspace{-0.25cm}
Prior works, including CLRQ\citep{clrq} (manifold geodesic distances for clustering), GLRSQ\citep{glrsq} (strategies for symmetric positive definite matrices), MANIQUEANT\citep{maniqueant} (FIM-based gradient mismatch mitigation), FIT\citep{fit} (information-geometric metrics for distortion reduction), and PLRQ\citep{plrq} (probabilistic vector quantization with manifold learning), have explored geometric approaches in quantization. However, these efforts remain fragmented—--each focuses on isolated geometric properties (e.g., matrix symmetry, manifold metrics) or individual quantization issues (e.g., gradient alignment, distortion control)—--without a unified framework to jointly optimize error direction and channel-wise bit allocation.

\vspace{-0.25cm}
\section{Preliminaries}\label{sec:preliminaries}
To facilitate the presentation of our proposed method, we first introduce three key components: Information Geometry and Riemannian Manifolds, Natural Gradient, and Cartesian Product Vector Quantization.
Specifically, these concepts form the theoretical foundation of our framework and provide a unified mathematical structure for subsequent sections.

\vspace{-0.25cm}
\subsection{Information Geometry and Riemannian Manifolds}\label{sec:preliminaries1}
\vspace{-0.25cm}
The parameter space of deep neural networks is often assumed to be a high-dimensional Euclidean space.
However, this assumption neglects the non-uniform sensitivity of different parameter directions.
Recent studies\citep{geometric, natural_gradient, riemannian, fisher_nature_gradient} demonstrate that the parameter space can be more accurately modeled as a Riemannian manifold endowed with the Fisher Information Metric\citep{fim}.
This geometric view better captures the sensitivity of the model to parameter perturbations, providing a theoretical basis for analyzing quantization errors.
Let $\mathcal{M}$ denote the differentiable manifold of network parameters $W$. \textbf{The Fisher Information Matrix(FIM)}, which we denote by $\mathbf{F}_W$, defines the metric tensor in the parameter space as:
\begin{equation}
    F_{ij}(W) = \mathbb{E}\!\left[\frac{\partial \log p(x|W)}{\partial W_i}\frac{\partial \log p(x|W)}{\partial W_j}\right],
\end{equation}
where $p(x|W)$ is the model's output distribution. This metric endows the parameter space with a Riemannian geometric structure, and the corresponding \textbf{inner product} is induced as follows:
\begin{equation}
    \langle \Delta W_1, \Delta W_2 \rangle_W = \Delta W_1^\top \mathbf{F}_W \Delta W_2.
\end{equation}
This inner product characterizes the geometric distance of parameter perturbations on the manifold, directly relating parameter changes to the KL divergence of the model output distribution (for the derivation process, see Appendix~\ref{sec:appendix_KL_FIM}). In practical applications, the channel-wise FIM sub-matrix $\mathbf{F}_c$ is used to approximate the global geometric structure for fine-grained sensitivity analysis.

\vspace{-0.25cm}
\subsection{Negative Natural Gradient} \label{sec:preliminaries2}
\vspace{-0.25cm}
In a flat Euclidean space, the standard negative gradient indicates the direction of the steepest descent. However, on a Riemannian manifold with complex curvature, the optimal descent direction is given by the \textbf{negative natural gradient}\citep{natural_gradient}:
\begin{equation}\label{eq:natural_gradient}
    -\tilde{\nabla} \mathcal{L} = -\mathbf{F}_W^{-1}\nabla \mathcal{L}.
\end{equation}
It geometrically corrects the Euclidean gradient through the FIM, ensuring that the parameter update descends along the shortest path (geodesic, defined in Appendix~\ref{sec:appendix_geodesic}) on the manifold. The negative natural gradient direction $-\tilde{\nabla}\mathcal{L}$ is the direction in which the loss function decreases the fastest, and it is also the "low-sensitivity direction" that the quantization error should try to align with, in order to minimize the impact of the error on the model performance. The derivation of the natural gradient can be based on the Taylor expansion and the Lagrange multiplier method. For the specific derivation steps, refer to Appendix~\ref{sec:appendix_natural_gradient}.

\vspace{-0.25cm}
\subsection{Product Vector Quantization}\label{sec:preliminaries_product_quantize}
\vspace{-0.25cm}
Product Quantization (PQ)\citep{product_quantize} extends vector quantization (VQ) by splitting high-dimensional vectors into sub-vectors and quantizing them independently, improving compression efficiency and scalability.
Similar to VQ, the optimization process of PQ involves finding the closest cluster center $\mathcal{C}_i$(the $i$-th codeword in the codebook) for an input vector $v$:
\begin{equation}\label{eq:kmeans}
    \arg\min_{i \in k} \left\| v - \mathcal{C}_i \right\|^2.
\end{equation}

For a detailed comparison between PQ and VQ, refer to Appendix~\ref{appendix:product_quantization}.
RSAVQ adopts this PQ formulation as its foundation and introduces two key enhancements: specifically, (1) error direction alignment via Riemannian projections, and (2) adaptive bit allocation guided by channel-wise curvature. These modifications jointly improve quantization accuracy under ultra-low-bit settings.
\vspace{-0.25cm}
\section{Method}\label{sec:method}
\vspace{-0.25cm}
Traditional quantization methods simplify optimization in Eq.~\ref{eq:kmeans} as a Euclidean problem, leading to suboptimal control of quantization error. Inspired by the Riemannian perspective in Preliminaries~\ref{sec:preliminaries1}---which models the parameter space (i.e. weights) as a curved manifold endowed with the Fisher information metric---we propose the Riemannian Sensitivity-Aware Vector Quantization (RSAVQ) framework. RSAVQ unifies information geometry and vector quantization via two tightly coupled modules: 
(1) \textbf{Error Direction Sensitivity Guidance(EDSG)}, which projects clustering errors onto low‑sensitivity (negative natural gradient) directions on the manifold; and  
(2) \textbf{Weight Channel Sensitivity Guidance(WCSG)}, which measures each channel’s local curvature to dynamically allocate bits and guide codebook updates.  
By integrating EDSG and WCSG, RSAVQ minimizes quantization distortion under extreme low-bit constraints while maintaining computational efficiency. 
For full algorithmic details, see Appendix~\ref{sec:appendix_algorithm1}.

\vspace{-0.25cm}
\subsection{EDSG: Error Direction Sensitivity Guidance}\label{sec:method1}
\vspace{-0.25cm}

During model quantization, quantization error inevitably occurs, defined as 
\begin{equation}
    E = W - \mathcal{C}(W)
\end{equation}
where $\mathcal{C}(\cdot)$ denotes the pseudo-quantization operator.
Traditional methods~\citep{vptq} typically rely on mean squared error (MSE) or other Euclidean-based metrics to quantify quantization error and determine optimal coefficients. However, these approaches overlook the geometric sensitivity associated with the direction of the error. Consequently, errors can amplify along directions that are highly sensitive to model performance, resulting in significant accuracy degradation. Although existing techniques~\citep{gptvq} attempt to mitigate such errors through post hoc compensation, their initial disregard for directional sensitivity fundamentally limits their effectiveness. To address this limitation, we propose an Error Projection algorithm based on information geometry. Specifically, we project the quantization error onto the negative natural gradient directions of the loss function, which correspond to low-sensitivity directions on the Riemannian manifold. This projection enables a more precise assessment of the quantization error’s impact on overall model performance.

\begin{wrapfigure}{r}{0.6\textwidth}
    \centering
    \vspace{-0.3cm}
    \includegraphics[width=\linewidth]{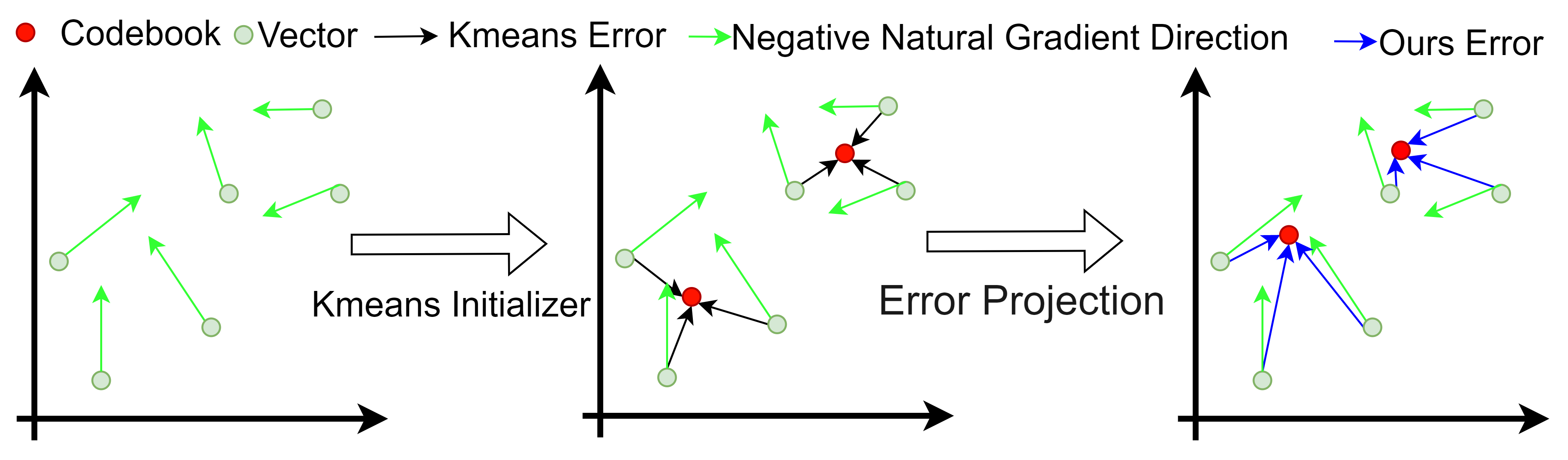}
    \vspace{-0.6cm}
    \caption{Clustering process of error projection along negative natural gradient direction.}
    \vspace{-0.6cm}
    \label{fig5:error_projection}
\end{wrapfigure}

As illustrated in Fig.~\ref{fig5:error_projection}, our goal is to project the quantization error onto directions that have minimal influence on the loss function, thereby reducing performance degradation from the perspective of information geometry.

Vector quantization (VQ) has been demonstrated to be a highly promising technique for large model compression\citep{vptq,qtip}, largely due to its inherent flexibility in adjusting the quantization error directions.
Building on this property, our objective is to design a codebook $C$ that encourages the quantization error $E$ to lie along "low-sensitivity" directions on the parameter manifold $\mathcal{M}$, thereby minimizing the growth of the KL divergence induced by quantization.
Through Lagrangian optimization (see Appendix~\ref{sec:appendix_natural_gradient}), we formally prove that projecting $E$ onto the negative natural gradient direction $-\tilde{\nabla} \mathcal{L} = -\mathbf{F}_W^{-1}\,\nabla \mathcal{L}$ minimizes the first-order increase in the loss function on the Riemannian manifold. Here, $F_W$ denotes the Fisher information matrix, characterizing the local geometry of the parameter space.
To enforce this projection during quantization, we define the negative natural gradient direction projection loss in the tangent space $T_W\mathcal{M}$ as:
\begin{equation}\label{eq:align_loss}
    \mathcal{L}_\text{project} = \|E + \lambda*\tilde{\nabla}\mathcal{L}\|_\mathbf{F}^2
\end{equation}
    where $\|\cdot\|_F$ denotes the norm induced by the Fisher metric, and $\lambda$ is a hyperparameter that controls the trade-off between strict error projection and quantization flexibility.
By minimizing $\mathcal{L}_\text{project}$, RSAVQ encourages the quantization error $E$ to remain in low-sensitivity directions, thus preserving model performance even under extremely low-bit quantization.

Building upon the product quantization (PQ) strategy introduced in Preliminaries~\ref{sec:preliminaries_product_quantize}, we decompose the weight vectors into a set of sub-vectors $\{v_i\}$, each independently quantized.
During the PQ optimization process, we incorporate the projection loss $\mathcal{L}_{\text{project}}$ in the negative natural gradient directions as an additional constraint, alternately optimizing both the codebooks and the assignment indices.
This procedure ensures that the quantization error is geometrically projected along the negative natural gradient direction, fundamentally minimizes loss increase and performance degradation
However, under a constrained bit budget, different channels exhibit varying tolerances to quantization error.
This observation motivates us to further exploit the local curvature information of the parameter manifold to dynamically allocate bit resources, thereby achieving a more fine-grained trade-off between accuracy and compression.
\vspace{-0.25cm}
\subsection{WCSG: Weight Channel Sensitivity Guidance}\label{sec:method2}
\vspace{-0.25cm}
\begin{figure}[!h]
    \centering
    \includegraphics[width=1.0\textwidth]{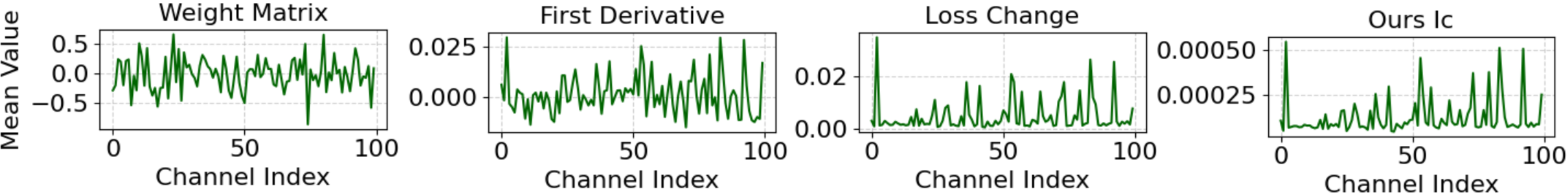}
    \caption{Weight channel sensitivity analysis and comparison.}
    \vspace{-0.25cm}
    \label{fig6:Ic_compare}
\end{figure}

Building on the channel sensitivity characterization via the Fisher Information Matrix, we propose a geometry-aware bit allocation strategy.
Our goal is to further minimize the global quantization distortion in addition to the error direction projection achieved in Method~\ref{sec:method1}.

In practice, different channels exhibit heterogeneous sensitivity to quantization perturbations.
Thus, a naive uniform bit allocation is suboptimal, and an adaptive strategy is needed to allocate bits proportionally to each channel's importance, thereby minimizing the overall loss degradation.

Existing Euclidean-based sensitivity metrics---such as the gradient magnitude used in SparseGPT\citep{sparsegpt} or the weight-activation product employed by Wanda\citep{wanda}---fail to capture the underlying curvature of the parameter space.
As demonstrated in Fig.~\ref{fig6:Ic_compare}, simple statistics like (a) the average weight magnitude or (b) the average gradient per channel show no clear correspondence with (c) the true loss sensitivity under perturbations.
This mismatch highlights the necessity of a curvature-aware measurement for accurate sensitivity estimation.

As introduced in Preliminaries~\ref{sec:preliminaries1}, the negative natural gradient is defined as $-\tilde{\nabla}\mathcal{L} = -\mathbf{F}_W^{-1} \nabla \mathcal{L}$.
In traditional Euclidean space, the negative gradient $-\nabla \mathcal{L}$ represents the direction of steepest descent. However, on a Riemannian manifold with non-uniform curvature, the optimal descent direction must be corrected by local metric $\mathbf{F}_W$, leading to the negative natural gradient formulation.

While the negative natural gradient addresses the geometrically optimal update direction, quantifying \emph{weight channel sensitivity} requires a scalar metric that integrates both gradient magnitude and local manifold curvature.
To this end, we leverage the Riemannian norm of the negative natural gradient, which measures how parameter perturbations in each channel's tangent space affect the loss function.
We decompose the weight tensor $W$ channel-wise as $\{W_c\}$, and for each channel $c$, we compute its Riemannian curvature energy $I_c$ (full derivation in Appendix~\ref{sec:appendix_Ic_cal}) as: \begin{equation} 
I_c = \frac{1}{2} |-\tilde{\nabla}\mathcal{L}_c|_W^2 = \frac{1}{2} (\tilde{\nabla}\mathcal{L}_c)^\top \mathbf{F}_c \tilde{\nabla}\mathcal{L}_c. \label{eq:Ic_cal} \end{equation}
Here, $\mathbf{F}_c$ is the FIM corresponding to channel $c$. The acquisition of FIM is approximated through Kronecker decomposition. For detailed acquisition steps and principles, refer to \ref{sec:appendix_fim}.
Intuitively, $I_c$ captures how sharply the loss landscape curves along the parameters of channel 
$c$: a larger $I_c$ indicates higher sensitivity to perturbations and thus a greater need for precise quantization.

Under a fixed total bit budget $B_{\max} = \sum b_c$, how to allocate bits across channels becomes crucial for minimizing global quantization distortion.
Based on the rate-distortion theory~\citep{rate_distortion}, the quantization distortion scales exponentially with the number of bits as $D\propto 2^{-2b}$.
Considering the channel curvature sensitivity $I_c$ as an amplification factor for distortion, the global distortion objective can be formulated as:
\begin{equation}
    Global \space Distortion=\sum I_c \cdot 2^{-2b_c}.
\end{equation}
Using Lagrangian optimization (details in Appendix~\ref{sec:appendix_bit_allocation}), the optimal bit allocation rule is derived as: $b_c \propto \log_2 I_c$, ensuring that more sensitive channels receive a greater bit allocation.

In practice, for a layer with $C$ channels, the bit assignment per channel is given by: 
\begin{equation} 
b_c = Round\left( B_{\max} \cdot \frac{\log_2 I_c}{\sum_{c=1}^{C} \log_2 I_c} \right). \label{eq:bit_cal} 
\end{equation} 

\begin{figure}[!t]
    \centering
    \vspace{-0.9cm}
    \includegraphics[width=0.80\textwidth]{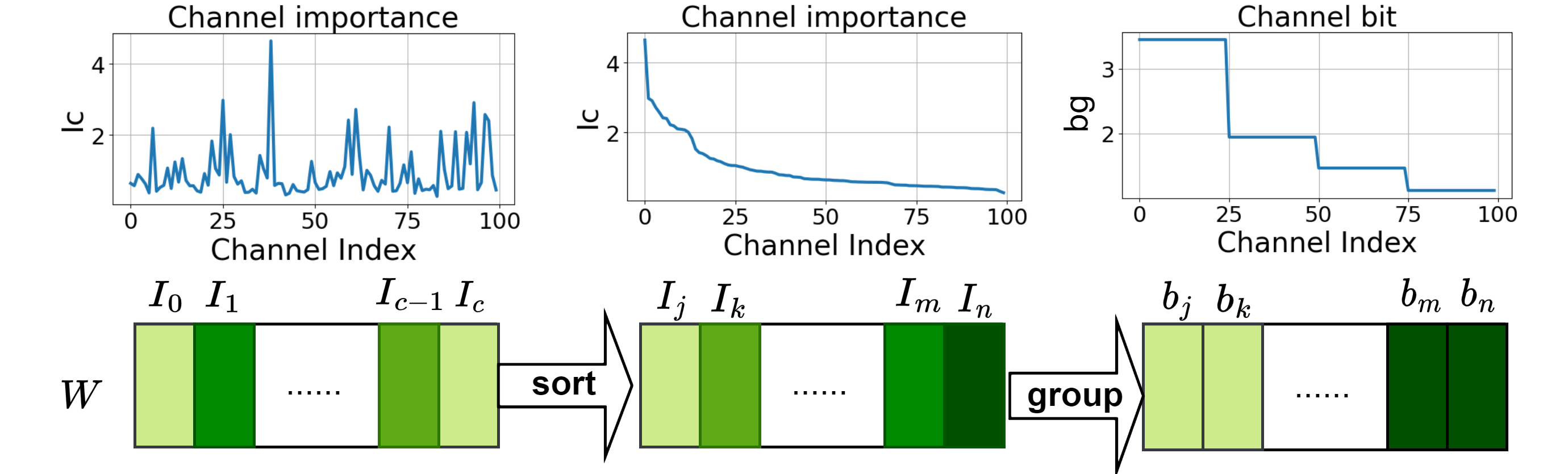}
    \caption{Channel sensitivity-driven channel grouping and bit assignment.}
    \vspace{-0.5cm}
    \label{fig7:bit_group}
\end{figure}

To enable codebook sharing (each group uses a single codebook) while preserving weights channel sensitivity, we introduce a sensitivity-ordered channel grouping strategy. 
As shown in Fig.~\ref{fig7:bit_group}. First, channels are sorted in descending order of their Riemannian curvature energy \(I_c\). We then partition them into \(G\) uniform groups, where the group size is \(n = \lceil C/G \rceil\) and the \(g\)-th group contains channels indexed by \(G_g = [(g-1)n+1, \min(gn, C)]\). For each group, a unified bit-width \(b_g\) is assigned by averaging the dynamically allocated bits of its members, that is,  
\begin{equation}\label{eq:bc_group_cal}
    b_g = \text{Round}\left(\frac{1}{|G_g|} \sum_{c \in G_g} b_c\right),
\end{equation}
where \(b_c \propto \log_2 I_c\) under the total bit budget constraint. This strategy clusters sensitive channels (high \(I_c\)) to receive more bits while aggressively compressing less sensitive groups, balancing distortion minimization with practical deployment constraints.

\vspace{1em}

\vspace{-0.6cm}
\section{Experiments}\label{experiments}
\vspace{-0.25cm}
\paragraph{Baseline.} 
We focus on weight-only quantization and compare RSAVQ with several strong PTQ baselines, including GPTQ\citep{gptq}, GPTVQ\citep{gptvq}, AQLM\citep{aqlm}, DB-LLM\citep{db-llm},QuIP\citep{quip}, QuIP\#\citep{quip_sharp}, and VPTQ\citep{vptq}.
Baseline results for these methods are cited from their original papers. For RSAVQ, we use a k-means-based VQ approach similar to VPTQ, with the following settings: the vector length is set to 6, and weight matrices are divided into 4 groups, with each sharing its own codebook.
Unless otherwise specified, all experiments are conducted on NVIDIA A100-80GB GPU.
\begin{table}[!th]
\centering
\small
\vspace{-0.3cm}
\caption{The perplexity (ppl) of various quantization algorithms on the LLaMA-2 models when the dataset is Wikitext-2 and the sequence length is 4096, as well as the test performance in various zero-shot tasks.}
\label{tab:llama2-models}
\resizebox{\textwidth}{!}{
\begin{tabular}{c|ccc|ccc|ccc}
\hline
       & \multicolumn{3}{c|}{LLaMA-2 7B}                     & \multicolumn{3}{c|}{LLaMA-2 13B}                    & \multicolumn{3}{c}{LLaMA-2 70B}            \\ \cline{2-10} 
\multirow{-2}{*}{Methods} &
  \multicolumn{1}{c|}{Bits} &
  W2↓ &
  0-shot Avg↑ &
  \multicolumn{1}{c|}{Bits} &
  W2↓ &
  0-shot Avg↑ &
  \multicolumn{1}{c|}{Bits} &
  W2↓ &
  0-shot Avg↑ \\ \hline
\rowcolor[HTML]{F2F3F5} 
FP16 &
  \multicolumn{1}{c|}{\cellcolor[HTML]{F2F3F5}16} &
  5.12 &
  64.7 &
  \multicolumn{1}{c|}{\cellcolor[HTML]{F2F3F5}16} &
  4.57 &
  67.82 &
  \multicolumn{1}{c|}{\cellcolor[HTML]{F2F3F5}16} &
  3.12 &
  70.21 \\
GPTQ   & \multicolumn{1}{c|}{2} & 50.75         & 39.16 & \multicolumn{1}{c|}{2} & 43.84         & 43.72 & \multicolumn{1}{c|}{2} & --   & 59.18 \\
GPTVQ  & \multicolumn{1}{c|}{2.25}  & 6.71          & 56.14 & \multicolumn{1}{c|}{2.25}  & 5.72          & 61.56 & \multicolumn{1}{c|}{2.25}  & 4.25 & 68.55 \\
DB-LLM & \multicolumn{1}{c|}{2.01}  & 7.23          & 55.12 & \multicolumn{1}{c|}{2.01}  & 6.19          & 59.41 & \multicolumn{1}{c|}{2.01}  & 4.64 & 65.83 \\
AQLM   & \multicolumn{1}{c|}{2.29}  & 6.29          & 58.57 & \multicolumn{1}{c|}{2.18}  & 5.41          & 61.58 & \multicolumn{1}{c|}{2.07}  & 3.94 & 68.75 \\
VPTQ   & \multicolumn{1}{c|}{2.02}  & 6.13          & 58.13 & \multicolumn{1}{c|}{2.02}  & 5.32          & 62.37 & \multicolumn{1}{c|}{2.07}  & 3.93 & 68.61 \\
QuIP\# & \multicolumn{1}{c|}{2}     & 6.19          & 58.22 & \multicolumn{1}{c|}{2}     & 5.35          & 61.96 & \multicolumn{1}{c|}{2}     & 3.91 & 68.94 \\
\rowcolor[HTML]{C0C0C0} 
RSAVQ &
  \multicolumn{1}{c|}{\cellcolor[HTML]{C0C0C0}2} &
  \textbf{5.97} &
  \textbf{58.66} &
  \multicolumn{1}{c|}{\cellcolor[HTML]{C0C0C0}2} &
  \textbf{5.29} &
  \textbf{62.84} &
  \multicolumn{1}{c|}{\cellcolor[HTML]{C0C0C0}2} &
  \textbf{3.55} &
  \textbf{69.05} \\ \hline
GPTQ   & \multicolumn{1}{c|}{3}     & 8.06          & 53.1  & \multicolumn{1}{c|}{3}     & 5.85          & 59.61 & \multicolumn{1}{c|}{3}     & 4.4  & 65.41 \\
GPTVQ  & \multicolumn{1}{c|}{3.125} & 5.44          & 62.69 & \multicolumn{1}{c|}{3.125} & 4.8           & 59.63 & \multicolumn{1}{c|}{3.125} & --   & --    \\
AQLM   & \multicolumn{1}{c|}{3.04}  & 5.46          & 60.88 & \multicolumn{1}{c|}{3.03}  & 4.82          & 63.49 & \multicolumn{1}{c|}{3.01}  & 3.36 & 69.86 \\
VPTQ   & \multicolumn{1}{c|}{3.02}  & 5.43          & 61.72 & \multicolumn{1}{c|}{3.03}  & 4.79          & 64.21 & \multicolumn{1}{c|}{3.01}  & 3.34 & 69.58 \\
QuIP\# & \multicolumn{1}{c|}{3}     & 5.41          & --    & \multicolumn{1}{c|}{3}     & 4.78          & --    & \multicolumn{1}{c|}{3}     & 3.35 & --    \\
\rowcolor[HTML]{C0C0C0} 
RSAVQ &
  \multicolumn{1}{c|}{\cellcolor[HTML]{C0C0C0}3.01} &
  \textbf{5.26} &
  \textbf{62.7} &
  \multicolumn{1}{c|}{\cellcolor[HTML]{C0C0C0}3.01} &
  \textbf{4.74} &
  \textbf{66.12} &
  \multicolumn{1}{c|}{\cellcolor[HTML]{C0C0C0}3.01} &
  \textbf{3.25} &
  \textbf{70.42} \\ \hline
GPTQ   & \multicolumn{1}{c|}{4}     & 5.49          & 60.64 & \multicolumn{1}{c|}{4}     & 4.78          & 63.87 & \multicolumn{1}{c|}{4}     & 3.35 & 69.25 \\
GPTVQ  & \multicolumn{1}{c|}{4.125} & 5.27          & 62.28 & \multicolumn{1}{c|}{4.125} & 5.27          & 64.28 & \multicolumn{1}{c|}{4.125} & --   & --    \\
AQLM   & \multicolumn{1}{c|}{4.04}  & 5.21          & 62.54 & \multicolumn{1}{c|}{3.94}  & 4.65          & 65.12 & \multicolumn{1}{c|}{4.14}  & 3.19 & 69.93 \\
VPTQ   & \multicolumn{1}{c|}{4.01}  & 5.26          & 61.98 & \multicolumn{1}{c|}{4.02}  & 4.64          & 64.89 & \multicolumn{1}{c|}{4.01}  & 3.19 & 69.8  \\
QuIP\# & \multicolumn{1}{c|}{4}     & \textbf{5.19} & --    & \multicolumn{1}{c|}{4}     & \textbf{4.63} & --    & \multicolumn{1}{c|}{4}     & 3.18 & --    \\
\rowcolor[HTML]{C0C0C0} 
RSAVQ &
  \multicolumn{1}{c|}{\cellcolor[HTML]{C0C0C0}4.01} &
  5.22 &
  \textbf{63.62} &
  \multicolumn{1}{c|}{\cellcolor[HTML]{C0C0C0}4.01} &
  4.72 &
  \textbf{66.82} &
  \multicolumn{1}{c|}{\cellcolor[HTML]{C0C0C0}4.01} &
  \textbf{3.11} &
  \textbf{70.34} \\ \hline
\end{tabular}
}
\vspace{-0.5cm}
\end{table}

\paragraph{Models and Datasets.}
We conduct experiments on LLaMA-2 7B, LLaMA-2 13B, LLaMA-2 70B\citep{llama2}, and LLaMA-3 8B, LLaMA-3 70B\citep{llama3} models to evaluate the performance of our proposed method. The calibration dataset used in our experiments is sampled from the Red\_Pajama dataset\citep{redpajama}.

To evaluate the performance of the baselines, we compute the perplexity (PPL) of the models on the WikiText-2 dataset\citep{wikitext2}. We evaluate the models by randomly sampling sequences from the dataset with the same length as the calibration data. Lower perplexity indicates better preservation of the original output distribution.
For direct comparison with methods like VPTQ\citep{vptq} and Quip\#\citep{quip_sharp}, we use the same sequence lengths during testing. Specifically, we test PPL with sequence lengths 4096 for the LLaMA-2 models and 2048 for the LLaMA-3 models.

Additionally, we evaluate generalization capability on several zero-shot tasks, including WinoGrand\citep{winogrande}, HellaSwag\citep{hellaswag}, PIQA\citep{piqa}, ARC-e\citep{arc-e}, and ARC-c\citep{arc-c}. All evaluations are performed using the open-source LM-Evaluation-Harness\citep{lm-eval} toolkit.

\begin{minipage}[t]{0.41\textwidth}
\paragraph{Main Experimental Results.}
Tab.~\ref{tab:llama2-models} presents the experimental results on the LLaMA-2 series models. We evaluate LLaMA-2 7B, LLaMA-2 13B, and LLaMA-2 70B under a 2-bit quantization configuration. For the LLaMA-2 7B model, RSAVQ achieves a perplexity of 5.97 at 2-bits, which is lower than VPTQ (6.13) and other methods, demonstrating more robust performance on WikiText-2. Furthermore, RSAVQ's zero-shot average accuracy is comparable to other methods and even exceeds them. For the LLaMA-2 70B model, in the 2-bit configuration, RSAVQ's performance is nearly identical to the FP16 baseline, with only a 0.4 PPL  
\end{minipage}
\hfill
\begin{minipage}[t]{0.55\textwidth}
\vspace{-0.8cm}
\begin{table}[H]
\caption{PPL on Wikitext-2(seq\_len=4096) and 0-shot task accuracy of PTQ algorithms on LLaMA-3.}
\label{tab:llama3-models}
\centering
\resizebox{\textwidth}{!}{
\begin{tabular}{c|ccc|ccc}
\hline
       & \multicolumn{3}{c|}{LLaMA-3 8B}           & \multicolumn{3}{c}{LLaMA-3 70B}           \\ \cline{2-7} 
\multirow{-2}{*}{Methods} &
  \multicolumn{1}{c|}{Bits} &
  W2↓ &
  0-shot Avg↑ &
  \multicolumn{1}{c|}{Bits} &
  W2↓ &
  0-shot Avg↑ \\ \hline
\rowcolor[HTML]{EFF0F1} 
FP16 &
  \multicolumn{1}{c|}{\cellcolor[HTML]{EFF0F1}16} &
  6.14 &
  68.66 &
  \multicolumn{1}{c|}{\cellcolor[HTML]{EFF0F1}16} &
  2.9 &
  75.32 \\
GPTQ   & \multicolumn{1}{c|}{2}    & 210  & 36.16 & \multicolumn{1}{c|}{2}    & 11.9 & 45.42 \\
QuIP   & \multicolumn{1}{c|}{2}    & 85.1 & 36.81 & \multicolumn{1}{c|}{2}    & 13   & 48.66 \\
QuIP\# & \multicolumn{1}{c|}{2}    & 9.11 & -- & \multicolumn{1}{c|}{2}   & 5.6   & --    \\
VPTQ   & \multicolumn{1}{c|}{2.08} & 9.29 & 60.22 & \multicolumn{1}{c|}{2.07} & 5.66 & 70.74 \\
\rowcolor[HTML]{C0C0C0} 
RSAVQ &
  \multicolumn{1}{c|}{\cellcolor[HTML]{C0C0C0}2} &
  8.79 &
  61.72 &
  \multicolumn{1}{c|}{\cellcolor[HTML]{C0C0C0}2} &
  5.6 &
  71.3 \\ \hline
GPTQ   & \multicolumn{1}{c|}{3}    & 8.2  & 61.7  & \multicolumn{1}{c|}{3}    & 5.2  & 70.58 \\
QuIP   & \multicolumn{1}{c|}{3}    & 7.5  & 63.72 & \multicolumn{1}{c|}{3}    & 4.7  & 72.56 \\
QuIP\# & \multicolumn{1}{c|}{3}    & 6.77 & -- & \multicolumn{1}{c|}{3}   & 3.8   & --    \\
VPTQ   & \multicolumn{1}{c|}{3.03} & 6.97 & 66.66 & \multicolumn{1}{c|}{3.01} & 3.81 & 73.68 \\
\rowcolor[HTML]{C0C0C0} 
RSAVQ &
  \multicolumn{1}{c|}{\cellcolor[HTML]{C0C0C0}3.01} &
  6.34 &
  66.38 &
  \multicolumn{1}{c|}{\cellcolor[HTML]{C0C0C0}3.01} &
  3.69 &
  74.26 \\ \hline
GPTQ   & \multicolumn{1}{c|}{4}    & 6.5  & 67.3  & \multicolumn{1}{c|}{4}    & 3.3  & 74.88 \\
QuIP   & \multicolumn{1}{c|}{4}    & 6.5  & 67.12 & \multicolumn{1}{c|}{4}    & 3.4  & 74.52 \\
QuIP\# & \multicolumn{1}{c|}{4}    & 6.34 & -- & \multicolumn{1}{c|}{4}   & 3.21   & --    \\
VPTQ   & \multicolumn{1}{c|}{4.03} & 6.42 & 68.14 & \multicolumn{1}{c|}{4.05} & 3.15 & 74.7  \\
\rowcolor[HTML]{C0C0C0} 
RSAVQ &
  \multicolumn{1}{c|}{\cellcolor[HTML]{C0C0C0}4.01} &
  6.31 &
  68.42 &
  \multicolumn{1}{c|}{\cellcolor[HTML]{C0C0C0}4.01} &
  3.11 &
  75.1 \\ \hline
\end{tabular}
}
\end{table}
\vspace{-0.3cm}
\end{minipage}
decrease and a 1.2 accuracy drop, indicating that the model can achieve near original accuracy with significantly reduced bitwidth. Tab.~\ref{tab:llama3-models} shows the results on LLaMA-3 8B and LLaMA-3 70B models. Under the 2 bit quantization configuration, RSAVQ achieves 61.72 and 71.3 zero-shot performance for the LLaMA-3 8B and LLaMA-3 70B models, respectively. Compared to other methods, such as GPTQ and VPTQ, RSAVQ not only maintains a lower perplexity but also ensures higher zero-shot task accuracy. These results validate the robustness of RSAVQ across different model scales.

\vspace{-0.4cm}
\paragraph{Ablation Study.}
To validate the importance of each component in RSAVQ, we performed ablation experiments to compare the performance of the k-means baseline method with the gradual addition of the two components: EDSG and WCSG.The test results are recorded in Tab.~\ref{tab:ablation_study_1}.
\begin{table}[!h]
\centering
\vspace{-0.45cm}
\caption{On the LLaMA-2 7B model, based on the Wikitext-2 dataset and 0-short datasets, test the effect comparison after using method 1 and method 2 respectively on the original Kmeans basis.}
\label{tab:ablation_study_1}
\begin{tabular}{cc|ccccccc}
\hline
\multicolumn{1}{c|}{} &          & \multicolumn{7}{c}{LLaMA-2 7B}                                        \\ \cline{3-9} 
\multicolumn{1}{c|}{\multirow{-2}{*}{Bits}} & \multirow{-2}{*}{Methods} & \multicolumn{1}{c|}{W2↓}                          & AC   & AE   & HE    & QA   & WI    & Acc Avg↑ \\ \hline
\rowcolor[HTML]{FFFFFF} 
\multicolumn{2}{c|}{FP16}& \multicolumn{1}{c|}{5.12} & 43.3 & 76.3 & 57.1  & 78.1 & 68.7  & 64.70   \\ \hline
\multicolumn{1}{c|}{} & Kmeans   & \multicolumn{1}{c|}{9.20}  & 28.9 & 62.5 & 43.3 & 71.5 & 63.3 & 53.90 \\
\multicolumn{1}{c|}{} & +EDSG & \multicolumn{1}{c|}{7.29 (\textcolor[HTML]{006400}{-1.91})} & 31.5 & 66.0 & 46.6 & 73.3 & 63.6 & 56.10 (\textcolor[HTML]{006400}{+2.20}) \\
\multicolumn{1}{c|}{\multirow{-3}{*}{2bit}} & +WCSG                  & \multicolumn{1}{c|}{5.81 (\textcolor[HTML]{006400}{-3.39})}                         & 37.2 & 64.4 & 50.7 & 75.4 & 65.7 & 58.69 (\textcolor[HTML]{006400}{+4.79})   \\ \hline
\multicolumn{1}{c|}{} & Kmeans   & \multicolumn{1}{c|}{7.25} & 35.0 & 68.6 & 47.7 & 73.4 & 64.6 & 57.86 \\
\multicolumn{1}{c|}{} & +EDSG & \multicolumn{1}{c|}{5.63 (\textcolor[HTML]{006400}{-1.62})} & 40.1 & 72.8 & 53.9 & 76.6 & 66.2 & 61.77 (\textcolor[HTML]{006400}{+3.91}) \\
\multicolumn{1}{c|}{\multirow{-3}{*}{3bit}} & +WCSG                  & \multicolumn{1}{c|}{5.26 (\textcolor[HTML]{006400}{-1.99})}                         & 41.0   & 73.0   & 54.7  & 76.7 & 68.2  & 62.70 (\textcolor[HTML]{006400}{+4.84})   \\ \hline
\end{tabular}
\vspace{-0.4cm}
\end{table}

On the LLaMA-2 7B model, ablation experiments on various datasets (AC, AE, HE, QA, WI) showed: In the 2-bit quantization setting, the k-means baseline achieved an average accuracy of 53.90. Adding the Error Direction Sensitivity Guidance (EDSG) module improved metrics such as AC,HE, thereby lifting average accuracy to 56.10. Introducing channel grouping further improved performance to 58.69. 
A similar trend in the 3-bit experiments confirmed that both modules reduce the quantization error and improve accuracy. The results validate their collaborative effect and the framework's effectiveness under extremely low-bit conditions.

\begin{wrapfigure}{r}{0.35\textwidth}
    \centering
    \includegraphics[width=\linewidth]{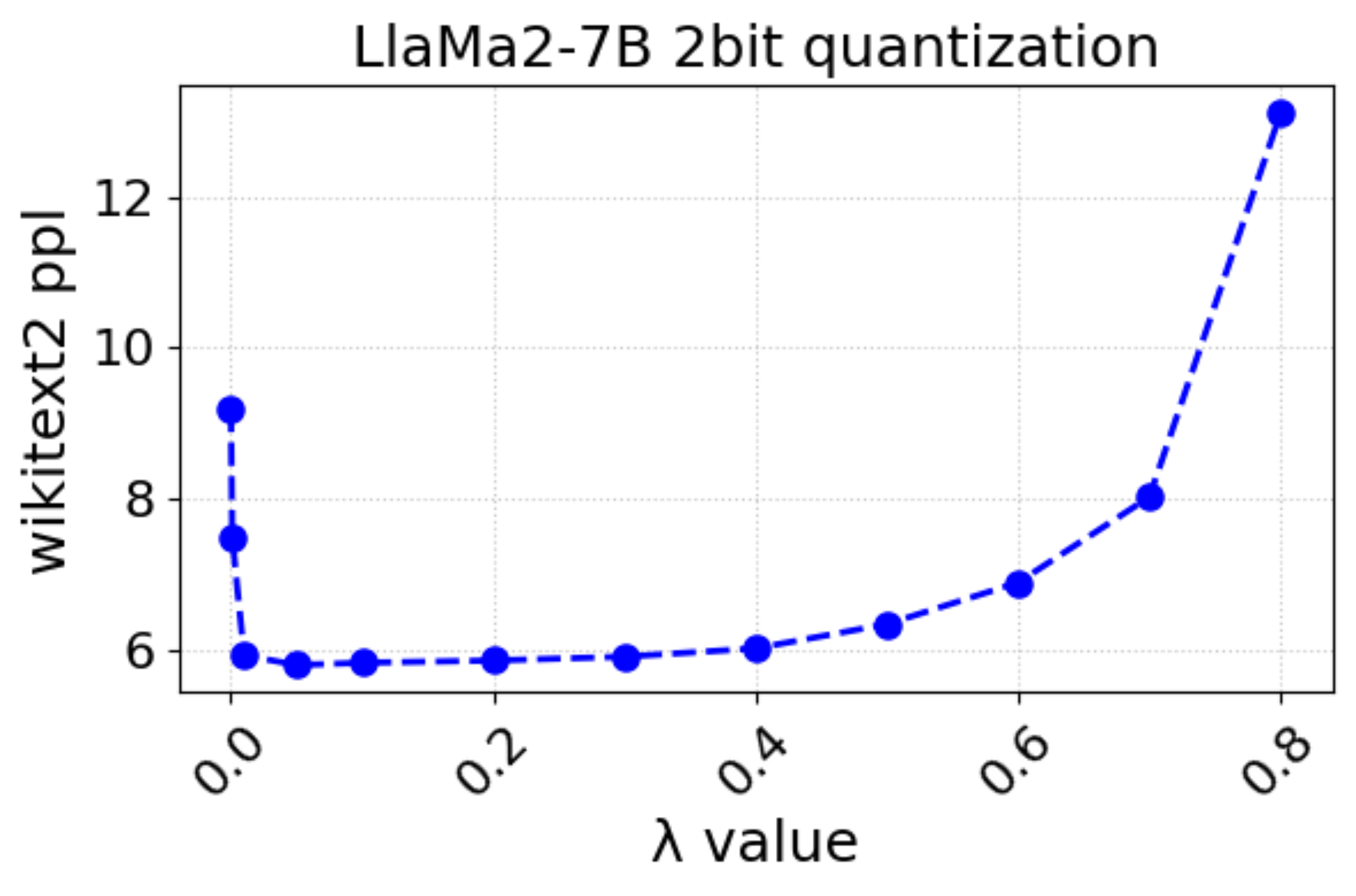}
    \caption{Optimal $\lambda$ Tuning in 2-bit quantized LLaMA-2 7B.}
    \label{fig8:lambda_ablation}
\end{wrapfigure}
We conducted an ablation experiment to evaluate the effect of the hyperparameter $\lambda$ on the performance of RSAVQ. Specifically, we focused on the LLaMA-2 7B model with 2-bit quantization and analyzed the PPL performance on the WikiText-2 dataset for various values of $\lambda$.
As shown in Fig.~\ref{fig8:lambda_ablation}. Our experiment showed that as $\lambda$ increased, the quantization accuracy first improved and then decreased. This indicates that $\lambda$ has an optimal range where the projection between quantization error and the natural gradient direction is most effective. 
Based on our experiments, we found that the optimal range for $\lambda$ lies between 0.01 and 0.1.Additional ablation studies across multiple models under 2-bit quantization are reported in Appendix~\ref{sec:appendix_lambda}, showing that $\lambda$ values in the range [0.01, 0.1] are robust across architectures.
This result highlights the importance of tuning $\lambda$ for balancing the projection strength between quantization error and natural gradient direction, with the optimal value providing the best trade-off between error reduction and model accuracy. We further analyze the sensitivity of RSAVQ to group size and codebook vector length in Appendix~\ref{sec:appendix_group_vector}.
\vspace{-0.3cm}
\paragraph{Speed and Efficiency Testing.}
In terms of hardware efficiency, we conducted speed and memory usage tests on the LLaMA-2 7B and LLaMA-2 13B models running inference on a single NVIDIA A100 GPU. The results demonstrate the significant improvements in inference speed and memory efficiency achieved by RSAVQ under low-bit quantization.

\begin{minipage}[t]{0.50\textwidth}
As shown in Tab.~\ref{tab:ablation_spped}, for the LLaMA-2 7B model, the inference speed with the FP16 model was 38.6 tokens/s, while under 2-bit quantization, the speed achieved a 1.57x speedup. Similarly, the LLaMA-2 13B model achieved comparable acceleration under 2-bit quantization. These results highlight 
\end{minipage}
\hfill
\begin{minipage}[t]{0.48\textwidth}
\vspace{-0.8cm}

\begin{table}[H]
\caption{Speed and Memory Testing}
\label{tab:ablation_spped}
\centering
\resizebox{\textwidth}{!}{
\begin{tabular}{c|cc|cc}
\hline
\multicolumn{1}{c|}{\multirow{2}{*}{Bits}} & \multicolumn{2}{c|}{LLaMA-2 7B}          & \multicolumn{2}{c}{LLaMA-2 13B}          \\ \cline{2-5} 
\multicolumn{1}{c|}{}                      & \multicolumn{1}{c|}{tokens/s} & memory(G) & \multicolumn{1}{c|}{tokens/s} & memory(G) \\ \hline
FP16 & \multicolumn{1}{c|}{38.60} & 13.16 & \multicolumn{1}{c|}{24.29} & 25.42 \\
2    & \multicolumn{1}{c|}{60.60} & 2.28  & \multicolumn{1}{c|}{44.21} & 4.03  \\
3    & \multicolumn{1}{c|}{50.57} & 3.83  & \multicolumn{1}{c|}{29.15} & 5.56  \\ \hline
\end{tabular}
}
\end{table}
\end{minipage}
that RSAVQ not only maintains low memory usage but also provides a significant speedup in inference, making it highly suitable for edge devices or resource-constrained environments where both memory and computation resources are limited. Beyond inference efficiency, RSAVQ also reduces the offline quantization time, as detailed in Appendix\ref{sec:appendix_quant_time}.

RSAVQ achieves superior low-bit performance by integrating information geometry and channel-wise bit allocation, outperforming traditional methods in PPL, zero-shot accuracy, and hardware efficiency across LLaMA models.

\vspace{-2mm}
\vspace{-0.25cm}
\section{Conclusion}
\vspace{-0.25cm}
This paper introduces RSAVQ, a Riemannian geometry-driven VQ framework that addresses extreme low-bit quantization challenges in LLMs. RSAVQ features two key innovations:
(1) Error Direction Sensitivity Guidance (EDSG), which projects quantization errors onto low-sensitivity directions (negative natural gradient) using the Fisher Information Matrix (FIM), minimizing performance degradation;
(2) Weight Channel Sensitivity Guidance (WCSG), which dynamically allocates bit resources based on FIM-derived curvature metrics to prioritize sensitive channels, balancing accuracy and compression efficiency.

Theoretically, RSAVQ bridges information geometry and neural network quantization by modeling the parameter space as a Riemannian manifold, enabling geometrically informed error control and adaptive bit allocation. The proposed Riemannian curvature energy metric $I_c$ offers a principled way to quantify channel sensitivity, overcoming Euclidean-based limitations. Practically, its hardware-friendly channel grouping strategy ensures efficient inference while achieving state-of-the-art 2-bit compression.

Experimental results on LLaMA models demonstrate RSAVQ’s superiority, particularly in extreme low-bit scenarios (e.g., 2-bit quantization with minimal PPL degradation and high zero-shot accuracy). By unifying geometric insights with quantization techniques, this work provides a robust solution for deploying LLMs in resource-constrained environments and opens new avenues for integrating information geometry into model optimization. While RSAVQ exhibits strong empirical performance, its limitations--—including limited cross-architecture and cross-domain generality, unexplored scalar quantization applicability, and hardware-specific memory efficiency constraints--—are analyzed in detail in Appendix~\ref{appendix:limitation}.

\vspace{-2mm}

\clearpage

{
\small
\bibliographystyle{plain}
\bibliography{neurips_2025}

\begin{thebibliography}{10}

\bibitem{llama4}
Meta AI.
\newblock Llama 4: Open foundation models for efficient multimodal understanding.
\newblock \url{https://llama.com/llama-downloads/}, 2025.
\newblock Available at: https://llama.com/llama-downloads/.

\bibitem{natural_gradient}
Shun-Ichi Amari.
\newblock Natural gradient works efficiently in learning.
\newblock {\em Neural computation}, 10(2):251--276, 1998.

\bibitem{fisher_nature_gradient}
Shun-ichi Amari, Ryo Karakida, and Masafumi Oizumi.
\newblock Fisher information and natural gradient learning in random deep networks.
\newblock In {\em The 22nd International Conference on Artificial Intelligence and Statistics}, pages 694--702. PMLR, 2019.

\bibitem{quarot}
Saleh Ashkboos, Amirkeivan Mohtashami, Maximilian Croci, Bo~Li, Pashmina Cameron, Martin Jaggi, Dan Alistarh, Torsten Hoefler, and James Hensman.
\newblock Quarot: Outlier-free 4-bit inference in rotated llms.
\newblock {\em Advances in Neural Information Processing Systems}, 37:100213--100240, 2024.

\bibitem{piqa}
Yonatan Bisk, Rowan Zellers, Jianfeng Gao, and Yejin Choi.
\newblock Piqa: Reasoning about physical commonsense in natural language.
\newblock In {\em Proceedings of the AAAI Conference on Artificial Intelligence}, volume~34, pages 7432--7439, 2020.

\bibitem{arc-e}
Michael Boratko, Harshit Padigela, Divyendra Mikkilineni, Pritish Yuvraj, Rajarshi Das, Andrew McCallum, Maria Chang, Achille Fokoue-Nkoutche, Pavan Kapanipathi, Nicholas Mattei, et~al.
\newblock A systematic classification of knowledge, reasoning, and context within the arc dataset.
\newblock {\em arXiv preprint arXiv:1806.00358}, 2018.

\bibitem{clrq}
Alice~Le Brigant and St{\'e}phane Puechmorel.
\newblock Optimal riemannian quantization with an application to air traffic analysis.
\newblock {\em arXiv preprint arXiv:1806.07605}, 2018.

\bibitem{geometric}
Michael~M Bronstein, Joan Bruna, Yann LeCun, Arthur Szlam, and Pierre Vandergheynst.
\newblock Geometric deep learning: going beyond euclidean data.
\newblock {\em IEEE Signal Processing Magazine}, 34(4):18--42, 2017.

\bibitem{quip}
Jerry Chee, Yaohui Cai, Volodymyr Kuleshov, and Christopher~M De~Sa.
\newblock Quip: 2-bit quantization of large language models with guarantees.
\newblock {\em Advances in Neural Information Processing Systems}, 36:4396--4429, 2023.

\bibitem{db-llm}
Hong Chen, Chengtao Lv, Liang Ding, Haotong Qin, Xiabin Zhou, Yifu Ding, Xuebo Liu, Min Zhang, Jinyang Guo, Xianglong Liu, et~al.
\newblock Db-llm: Accurate dual-binarization for efficient llms.
\newblock {\em arXiv preprint arXiv:2402.11960}, 2024.

\bibitem{maniqueant}
Jun Chen, Hanwen Chen, Jiangning Zhang, Tianxin Huang, Yong Liu, et~al.
\newblock Riemannian manifold embeddings for straight-through estimator.
\newblock {\em OpenReview}, 2022.

\bibitem{geodesic_distance}
Elwin~Bruno Christoffel.
\newblock Ueber die transformation der homogenen differentialausdr{\"u}cke zweiten grades.
\newblock {\em Journal für die reine und angewandte Mathematik}, 1869.

\bibitem{arc-c}
Peter Clark, Isaac Cowhey, Oren Etzioni, Tushar Khot, Ashish Sabharwal, Carissa Schoenick, and Oyvind Tafjord.
\newblock Think you have solved question answering? try arc, the ai2 reasoning challenge.
\newblock {\em arXiv preprint arXiv:1803.05457}, 2018.

\bibitem{qwen}
Alibaba Cloud.
\newblock Qwen: A versatile large language model for multimodal understanding.
\newblock \url{https://modelscope.cn/models/qwen}, 2025.
\newblock Available at: https://modelscope.cn/models/qwen.

\bibitem{aqlm}
Vage Egiazarian, Andrei Panferov, Denis Kuznedelev, Elias Frantar, Artem Babenko, and Dan Alistarh.
\newblock Extreme compression of large language models via additive quantization.
\newblock {\em arXiv preprint arXiv:2401.06118}, 2024.

\bibitem{geometric_survey}
Yanhong Fei, Yingjie Liu, Chentao Jia, Zhengyu Li, Xian Wei, and Mingsong Chen.
\newblock A survey of geometric optimization for deep learning: from euclidean space to riemannian manifold.
\newblock {\em ACM Computing Surveys}, 57(5):1--37, 2025.

\bibitem{sparsegpt}
Elias Frantar and Dan Alistarh.
\newblock Sparsegpt: Massive language models can be accurately pruned in one-shot.
\newblock In {\em International Conference on Machine Learning}, pages 10323--10337. PMLR, 2023.

\bibitem{gptq}
Elias Frantar, Saleh Ashkboos, Torsten Hoefler, and Dan Alistarh.
\newblock Gptq: Accurate post-training quantization for generative pre-trained transformers.
\newblock {\em arXiv preprint arXiv:2210.17323}, 2022.

\bibitem{lm-eval}
Leo Gao, Jonathan Tow, Stella Biderman, Sid Black, Anthony DiPofi, Charles Foster, Laurence Golding, Jeffrey Hsu, Kyle McDonell, and Niklas Muennighoff.
\newblock A framework for few-shot language model evaluation.
\newblock Technical report, GitHub, September 2021.

\bibitem{vector_quantize}
Allen Gersho.
\newblock Asymptotically optimal block quantization.
\newblock {\em IEEE Transactions on information theory}, 25(4):373--380, 1979.

\bibitem{rate_distortion}
Allen Gersho and Robert~M Gray.
\newblock {\em Vector quantization and signal compression}, volume 159.
\newblock Springer Science \& Business Media, 2012.

\bibitem{moequant}
Xing Hu, Zhixuan Chen, Dawei Yang, Zukang Xu, Chen Xu, Zhihang Yuan, Sifan Zhou, and Jiangyong Yu.
\newblock Moequant: Enhancing quantization for mixture-of-experts large language models via expert-balanced sampling and affinity guidance.
\newblock {\em arXiv preprint arXiv:2505.03804}, 2025.

\bibitem{ostquant}
Xing Hu, Yuan Cheng, Dawei Yang, Zukang Xu, Zhihang Yuan, Jiangyong Yu, Chen Xu, Zhe Jiang, and Sifan Zhou.
\newblock Ostquant: Refining large language model quantization with orthogonal and scaling transformations for better distribution fitting.
\newblock {\em arXiv preprint arXiv:2501.13987}, 2025.

\bibitem{illm}
Xing Hu, Yuan Cheng, Dawei Yang, Zhihang Yuan, Jiangyong Yu, Chen Xu, and Sifan Zhou.
\newblock I-llm: Efficient integer-only inference for fully-quantized low-bit large language models.
\newblock {\em arXiv preprint arXiv:2405.17849}, 2024.

\bibitem{ptq}
Benoit Jacob, Skirmantas Kligys, Bo~Chen, Menglong Zhu, Matthew Tang, Andrew Howard, Hartwig Adam, and Dmitry Kalenichenko.
\newblock Quantization and training of neural networks for efficient integer-arithmetic-only inference.
\newblock In {\em Proceedings of the IEEE conference on computer vision and pattern recognition}, pages 2704--2713, 2018.

\bibitem{product_quantize}
Herve Jegou, Matthijs Douze, and Cordelia Schmid.
\newblock Product quantization for nearest neighbor search.
\newblock {\em IEEE transactions on pattern analysis and machine intelligence}, 33(1):117--128, 2010.

\bibitem{riemannian}
Isay Katsman, Eric Chen, Sidhanth Holalkere, Anna Asch, Aaron Lou, Ser~Nam Lim, and Christopher~M De~Sa.
\newblock Riemannian residual neural networks.
\newblock {\em Advances in Neural Information Processing Systems}, 36:63502--63514, 2023.

\bibitem{2016pruning}
Hao Li, Asim Kadav, Igor Durdanovic, Hanan Samet, and Hans~Peter Graf.
\newblock Pruning filters for efficient convnets.
\newblock In {\em International Conference on Learning Representations (ICLR)}, 2017.
\newblock arXiv preprint arXiv:1608.08710.

\bibitem{awq}
Ji~Lin, Jiaming Tang, Haotian Tang, Shang Yang, Wei-Ming Chen, Wei-Chen Wang, Guangxuan Xiao, Xingyu Dang, Chuang Gan, and Song Han.
\newblock Awq: Activation-aware weight quantization for on-device llm compression and acceleration.
\newblock {\em Proceedings of Machine Learning and Systems}, 6:87--100, 2024.

\bibitem{llava}
Haotian Liu, Chunyuan Li, Qingyang Wu, and Yong~Jae Lee.
\newblock Visual instruction tuning.
\newblock {\em Advances in neural information processing systems}, 36:34892--34916, 2023.

\bibitem{liu2020deep}
Xue Liu, Weijie Xia, and Zhimiao Fan.
\newblock A deep neural network pruning method based on gradient l1-norm.
\newblock In {\em 2020 IEEE 6th International Conference on Computer and Communications (ICCC)}, pages 2070--2074. IEEE, 2020.

\bibitem{vptq}
Yifei Liu, Jicheng Wen, Yang Wang, Shengyu Ye, Li~Lyna Zhang, Ting Cao, Cheng Li, and Mao Yang.
\newblock Vptq: Extreme low-bit vector post-training quantization for large language models.
\newblock {\em arXiv preprint arXiv:2409.17066}, 2024.

\bibitem{natural_gradient_descent}
James Martens.
\newblock New insights and perspectives on the natural gradient method.
\newblock {\em Journal of Machine Learning Research}, 21(146):1--76, 2020.

\bibitem{k-fac}
James Martens and Roger Grosse.
\newblock Optimizing neural networks with kronecker-factored approximate curvature.
\newblock In {\em International conference on machine learning}, pages 2408--2417. PMLR, 2015.

\bibitem{wikitext2}
Stephen Merity, Caiming Xiong, James Bradbury, and Richard Socher.
\newblock Pointer sentinel mixture models.
\newblock {\em arXiv preprint arXiv:1609.07843}, 2016.

\bibitem{llama3}
AI~Meta.
\newblock Introducing meta llama 3: The most capable openly available llm to date.
\newblock {\em Meta AI}, 2(5):6, 2024.

\bibitem{drive}
Dhananjay Saikumar and Blesson Varghese.
\newblock Drive: Dual gradient-based rapid iterative pruning.
\newblock {\em arXiv preprint arXiv:2404.03687}, 2024.

\bibitem{winogrande}
Keisuke Sakaguchi, Ronan~Le Bras, Chandra Bhagavatula, and Yejin Choi.
\newblock Winogrande: An adversarial winograd schema challenge at scale.
\newblock {\em Communications of the ACM}, 64(9):99--106, 2021.

\bibitem{wanda}
Mingjie Sun, Zhuang Liu, Anna Bair, and J~Zico Kolter.
\newblock A simple and effective pruning approach for large language models.
\newblock {\em arXiv preprint arXiv:2306.11695}, 2023.

\bibitem{glrsq}
Fengzhen Tang, Mengling Fan, and Peter Ti{\v{n}}o.
\newblock Generalized learning riemannian space quantization: A case study on riemannian manifold of spd matrices.
\newblock {\em IEEE Transactions on Neural Networks and Learning Systems}, 32(1):281--292, 2020.

\bibitem{plrq}
Fengzhen Tang, Haifeng Feng, Peter Tino, Bailu Si, and Daxiong Ji.
\newblock Probabilistic learning vector quantization on manifold of symmetric positive definite matrices.
\newblock {\em Neural Networks}, 142:105--118, 2021.

\bibitem{redpajama}
{Together Computer}.
\newblock Redpajama-data-1t: An open dataset for training large language models.
\newblock \url{https://huggingface.co/datasets/togethercomputer/RedPajama-Data-1T}, 2023.
\newblock Available on Hugging Face Datasets.

\bibitem{llama2}
Hugo Touvron, Louis Martin, Kevin Stone, Peter Albert, Amjad Almahairi, Yasmine Babaei, Nikolay Bashlykov, Soumya Batra, Prajjwal Bhargava, Shruti Bhosale, et~al.
\newblock Llama 2: Open foundation and fine-tuned chat models.
\newblock {\em arXiv preprint arXiv:2307.09288}, 2023.

\bibitem{quip_sharp}
Albert Tseng, Jerry Chee, Qingyao Sun, Volodymyr Kuleshov, and Christopher De~Sa.
\newblock Quip\#: Even better llm quantization with hadamard incoherence and lattice codebooks.
\newblock {\em arXiv preprint arXiv:2402.04396}, 2024.

\bibitem{qtip}
Albert Tseng, Qingyao Sun, David Hou, and Christopher~M De~Sa.
\newblock Qtip: Quantization with trellises and incoherence processing.
\newblock {\em Advances in Neural Information Processing Systems}, 37:59597--59620, 2024.

\bibitem{gptvq}
Mart Van~Baalen, Andrey Kuzmin, Markus Nagel, Peter Couperus, Cedric Bastoul, Eric Mahurin, Tijmen Blankevoort, and Paul Whatmough.
\newblock Gptvq: The blessing of dimensionality for llm quantization.
\newblock {\em arXiv preprint arXiv:2402.15319}, 2024.

\bibitem{fim}
DA~Wagenaar.
\newblock Information geometry for neural networks.
\newblock {\em Term paper for reading course with ACC Coolen, King’s College London}, 1998.

\bibitem{rwkvquant}
Chen Xu, Yuxuan Yue, Zukang Xu, Xing Hu, Jiangyong Yu, Zhixuan Chen, Sifan Zhou, Zhihang Yuan, and Dawei Yang.
\newblock Rwkvquant: Quantizing the rwkv family with proxy guided hybrid of scalar and vector quantization.
\newblock {\em arXiv preprint arXiv:2505.03803}, 2025.

\bibitem{crvq}
Yuzhuang Xu, Shiyu Ji, Qingfu Zhu, and Wanxiang Che.
\newblock Crvq: Channel-relaxed vector quantization for extreme compression of llms.
\newblock {\em arXiv preprint arXiv:2412.09282}, 2024.

\bibitem{mambaquant}
Zukang Xu, Yuxuan Yue, Xing Hu, Zhihang Yuan, Zixu Jiang, Zhixuan Chen, Jiangyong Yu, Chen Xu, Sifan Zhou, and Dawei Yang.
\newblock Mambaquant: Quantizing the mamba family with variance aligned rotation methods.
\newblock {\em arXiv preprint arXiv:2501.13484}, 2025.

\bibitem{pcdvq}
Yuxuan Yue, Zukang Xu, Zhihang Yuan, Dawei Yang, Jianlong Wu, and Liqiang Nie.
\newblock Pcdvq: Enhancing vector quantization for large language models via polar coordinate decoupling.
\newblock {\em arXiv preprint arXiv:2506.05432}, 2025.

\bibitem{fit}
Ben Zandonati, Adrian~Alan Pol, Maurizio Pierini, Olya Sirkin, and Tal Kopetz.
\newblock Fit: A metric for model sensitivity.
\newblock {\em arXiv preprint arXiv:2210.08502}, 2022.

\bibitem{hellaswag}
Rowan Zellers, Ari Holtzman, Yonatan Bisk, Ali Farhadi, and Yejin Choi.
\newblock Hellaswag: Can a machine really finish your sentence?
\newblock {\em arXiv preprint arXiv:1905.07830}, 2019.

\bibitem{gps}
Zhi Zhang, Qizhe Zhang, Zijun Gao, Renrui Zhang, Ekaterina Shutova, Shiji Zhou, and Shanghang Zhang.
\newblock Gradient-based parameter selection for efficient fine-tuning.
\newblock In {\em Proceedings of the IEEE/CVF Conference on Computer Vision and Pattern Recognition}, pages 28566--28577, 2024.

\end{thebibliography}
}

\clearpage
\appendix
\section{Appendix}
\subsection{Limitations}\label{appendix:limitation}

Despite its strong empirical gains, RSAVQ has three main limitations that merit further study:  

\begin{itemize}  
    \item \textbf{Domain and architecture generality.} RSAVQ is validated primarily on Transformer-based language models (e.g., LLaMA), with limited empirical exploration in multimodal models, computer vision architectures, or reinforcement learning frameworks.  
    \item \textbf{Quantization method scope.} This work focuses exclusively on applying information geometry to vector quantization (VQ), while its potential utility in scalar quantization (SQ) remains unexplored. Extending geometric sensitivity analysis to SQ could reveal new optimization strategies for low-bit quantization.  
    \item \textbf{Hardware deployment efficiency.} Vector quantization’s reliance on codebook lookups introduces memory access overhead, particularly on heterogeneous hardware (CPUs/GPUs/edge devices), with limited optimization for platform-specific memory systems (e.g., cache-friendly indexing, parallel computation).  
\end{itemize}  

Future work will explore cross-architecture generalization, scalar quantization extensions, and hardware-aware quantization optimizations to enhance RSAVQ’s practical applicability.

\subsection{Acknowledgments}\label{sec:appendix_ack}
We would like to acknowledge the assistance of large language models in improving the clarity and readability of the manuscript text. We emphasize that all experiments, analyses, and core methodological innovations presented in this paper were independently conceived and executed by the authors, without the use of large language models for generation. 

We are also grateful to the reviewers, the area chairs and the program chairs for their valuable time and constructive feedback during the review process, which helped us significantly improve the quality of this work.

\subsection{Vector Quantization vs. Product Quantization}\label{appendix:product_quantization}
\begin{figure}[ht]
    \centering
    \includegraphics[width=1.0\textwidth]{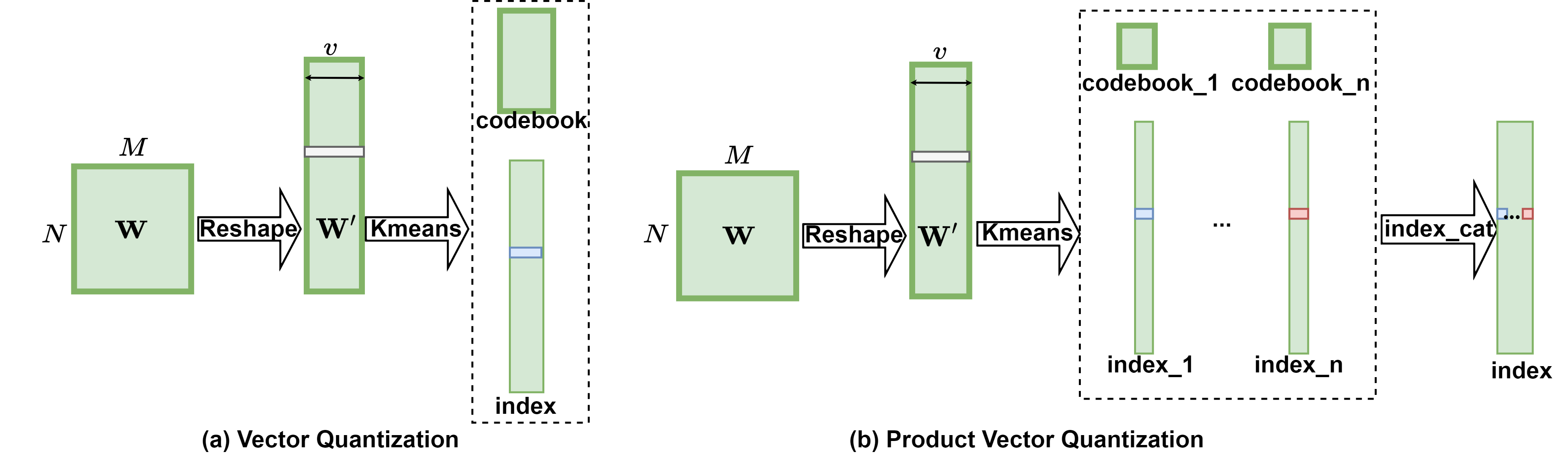}
    \caption{Schematic diagram of standard vector quantization (a) and product vector quantization (b).}
    \label{fig4:product_quantize_appendix}
\end{figure}

\paragraph{Vector Quantization (VQ).}
As shown in Fig.~\ref{fig4:product_quantize_appendix}(a), given a weight matrix $W \in \mathbb{R}^{M \times N}$, we first reshape it into a sequence of $v$-dimensional vectors $W' \in \mathbb{R}^{\frac{MN}{v} \times v}$. A codebook $\mathcal{C} \in \mathbb{R}^{k \times v}$ is then built using $k$-means clustering.

Each vector in $W'$ is quantized by selecting the nearest centroid using:
\begin{equation}
    \arg\min_{i \in \{1,\dots,k\}} \left\| v - \mathcal{C}_i \right\|^2,
\end{equation}
where $\mathcal{C}_i$ denotes the $i$-th codeword. The quantized results are stored as an index array referencing the codebook entries.

\paragraph{Product Quantization (PQ).}
To scale quantization to high-dimensional vectors, PQ divides each $v$-dimensional vector into $n$ equal-length sub-vectors:
\begin{equation}
v = [v_1, v_2, \dots, v_n], \quad v_i \in \mathbb{R}^{v/n}.
\end{equation}
Each sub-vector $v_i$ is independently quantized using a dedicated codebook $\mathcal{C}_i \in \mathbb{R}^{k \times v/n}$, producing $n$ sub-indices. These indices are concatenated into a single composite index that jointly encodes the original vector.

\subsection{Principle of Negative Natural Gradient}\label{sec:appendix_natural_gradient}
Referring to\citep{natural_gradient_descent}, the derivation process is as follows:

Let the small change of the loss function $L(W)$ at point $W$ be $\Delta L$. According to the first-order Taylor expansion, we have: 
\begin{equation}
    \Delta L\approx\nabla L^{\top}\Delta W
\end{equation}
On the Riemannian manifold, we hope to find a direction $\Delta W$ so that the loss function decreases the fastest under the given Riemannian metric. According to the Lagrange multiplier method, introduce the constraint condition $\langle\Delta W,\Delta W\rangle_{W}=1$, and construct the Lagrange function: 
\begin{equation}
    L(\Delta W,\lambda)=\nabla L^{\top}\Delta W-\lambda(\Delta W^{\top}F_{W}\Delta W - 1)
\end{equation}
Take the partial derivative of $\Delta W$ and set it to zero to obtain:
\begin{equation}
    \nabla L - 2\lambda F_{W}\Delta W = 0
\end{equation}
Get $\Delta W=\frac{2}{\lambda}F_{W}^{-1}\nabla L$. Let $-\widetilde{\nabla}L = -F_{W}^{-1}\nabla L$, which is the negative natural gradient.

\subsection{Detailed Derivation of KL Divergence Approximation  }\label{sec:appendix_KL_FIM}
The following presents the detailed derivation of the KL divergence approximation formula, involving Taylor expansion, the definition of the Fisher Information Matrix (FIM), and differential properties of probabilistic models:  

For parameterized probability distributions \( p(x|W) \) and \( p(x|W+\Delta W) \), the KL divergence is defined as:  
\begin{equation}
D_{\text{KL}}(p(x|W) \,\|\, p(x|W+\Delta W)) = \mathbb{E}_{x \sim p(x|W)} \left[ \log \frac{p(x|W)}{p(x|W+\Delta W)} \right].
\end{equation}  
Expanding the logarithmic term:  
\begin{equation}
D_{\text{KL}} = \mathbb{E} \left[ \log p(x|W) - \log p(x|W+\Delta W) \right].
\end{equation}  

Let \( f(W) = \log p(x|W) \). For a sufficiently small parameter perturbation \( \Delta W \), the second-order Taylor expansion of \( f(W+\Delta W) \) around \( W \) (neglecting higher-order terms) is:  
\begin{equation}
f(W+\Delta W) \approx f(W) + \nabla f(W)^\top \Delta W + \frac{1}{2} \Delta W^\top \mathbf{H}_f(W) \Delta W,
\end{equation}  
where:  
- \( \nabla f(W) = \frac{\partial \log p(x|W)}{\partial W} \) is the gradient (score function),  
- \( \mathbf{H}_f(W) = \frac{\partial^2 \log p(x|W)}{\partial W \partial W^\top} \) is the Hessian matrix.  

Substituting into the KL divergence expression:  
\begin{equation}
D_{\text{KL}} \approx \mathbb{E} \left[ f(W) - \left( f(W) + \nabla f^\top \Delta W + \frac{1}{2} \Delta W^\top \mathbf{H}_f \Delta W \right) \right] = \mathbb{E} \left[ -\nabla f^\top \Delta W - \frac{1}{2} \Delta W^\top \mathbf{H}_f \Delta W \right].
\end{equation}  

By the \textbf{property of the score function} (regularity condition of probability distributions):  
\begin{equation}
\mathbb{E} \left[ \nabla \log p(x|W) \right] = \int p(x|W) \cdot \nabla \log p(x|W) \, dx = \nabla \int p(x|W) \, dx = \nabla 1 = \mathbf{0}.
\end{equation}  
Thus, the expectation of the first-order term vanishes:  
\begin{equation}
\mathbb{E} \left[ -\nabla f^\top \Delta W \right] = -\Delta W^\top \mathbb{E} \left[ \nabla f \right] = \mathbf{0}.
\end{equation}  

Using the \textbf{Information Matrix Equality}:  
\begin{equation}
\mathbb{E} \left[ \mathbf{H}_f(W) \right] = -\mathbb{E} \left[ \nabla \log p(x|W) \nabla \log p(x|W)^\top \right] = -\mathbf{F}_W,
\end{equation}  
where \( \mathbf{F}_W \) is the Fisher Information Matrix (FIM), defined as:  
\begin{equation}
\mathbf{F}_W = \mathbb{E} \left[ \nabla \log p(x|W) \nabla \log p(x|W)^\top \right].
\end{equation}  
The expectation of the second-order term becomes:  
\begin{equation}
\mathbb{E} \left[ -\frac{1}{2} \Delta W^\top \mathbf{H}_f \Delta W \right] = \frac{1}{2} \Delta W^\top \mathbb{E} \left[ -\mathbf{H}_f \right] \Delta W = \frac{1}{2} \Delta W^\top \mathbf{F}_W \Delta W.
\end{equation}  

Under the Riemannian metric defined in the paper, the inner product is:  
\begin{equation}
\langle \Delta W_1, \Delta W_2 \rangle_W = \Delta W_1^\top \mathbf{F}_W \Delta W_2,
\end{equation}  
so the second-order term can be expressed as:  
\begin{equation}
\frac{1}{2} \Delta W^\top \mathbf{F}_W \Delta W = \frac{1}{2} \langle \Delta W, \Delta W \rangle_W.
\end{equation}  

Neglecting higher-order infinitesimal terms, the KL divergence approximates to:  
\begin{equation}
D_{\text{KL}}(p(x|W) \,\|\, p(x|W+\Delta W)) \approx \frac{1}{2} \langle \Delta W, \Delta W \rangle_W.
\end{equation}  
This approximation reveals the direct relationship between parameter perturbations in the Riemannian manifold (measured by FIM) and changes in the model's output distribution (KL divergence).  

In practical applications, to reduce computational overhead or enable fine-grained analysis, block-wise approximations (such as diagonal blocks or K-FAC approximation\citep{k-fac}) are often used to decompose the global Fisher matrix into channel-level (or subnetwork-level) submatrices \( \mathbf{F}_c \). These submatrices provide local geometric information for each channel, laying the foundation for subsequent sensitivity analysis and bit allocation.

\subsection{Geodesic}\label{sec:appendix_geodesic}
In Riemannian manifolds, the shortest path between two points is called a geodesic\citep{geodesic_distance}, and its length is defined as the geodesic distance. For two points $p,q\in M$ on the manifold, their \textbf{geodesic distance} definition is:
\begin{equation}
    d(p, q) = \int_0^1 \|\dot{\gamma}(t)\| \, dt.
\end{equation}
Here, $\gamma(t)$ is the geodesic that connects $p$ and $q$, and $\dot{\gamma}(t)$ is the geodesic velocity vector.
Geodesic distance is an important measure of the relationship between points on a manifold. Especially in quantitative error analysis, it can help us evaluate the impact of quantization error on model performance. For example, in our method, the direction of quantization error is projected onto a low-sensitivity direction to minimize its contribution to the geodesic distance.
If the manifold is compared to the surface of the earth, then the geodesic distance is the length of the great circular arc between two points. In the parameter space of the neural network, the geodesic distance on the manifold based on the $F_W$ measure in Preliminaries~\ref{sec:preliminaries1} reflects the actual impact of the parameter changes on the performance of the model.

\subsection{Riemannian Curvature Energy}\label{sec:appendix_Ic_cal}
For each channel \( c \), we aim to measure the impact of parameter perturbations on the loss function. In deep learning, methods relying on the magnitude of weight gradients to determine weight importance have long dominated, as the gradient magnitude directly reflects the dynamic contribution of weights to the loss function. Early studies \citep{2016pruning} proposed using the L2 norm of weights to assess importance, but subsequent work revealed that gradient information can more accurately capture the real-time impact of weights. For example, \citep{liu2020deep,gps} introduced the use of the L1 norm of gradients to filter important weights, and \citep{drive} further incorporated gradient direction information by fusing L1 and L2 norms for weight scoring. However, directly using the Euclidean norm of gradients \( \|\nabla \mathcal{L}_c\| \) overlooks a critical fact: the parameter space of deep neural networks is inherently a Riemannian manifold. Therefore, instead of directly using gradients, we employ the Riemannian norm of the negative natural gradient:  
\begin{equation}
    \|-\tilde{\nabla}\mathcal{L}_c\|_W^2 = \tilde{\nabla}\mathcal{L}_c^\top \mathbf{F}_c \tilde{\nabla}\mathcal{L}_c,
\end{equation}
where the negative natural gradient \( -\tilde{\nabla}\mathcal{L}_c = -\mathbf{F}_c^{-1} \nabla \mathcal{L}_c \) projects the Euclidean gradient onto the tangent space of the Riemannian manifold. Substituting the negative natural gradient into the norm yields:  
\begin{equation}
    \|-\tilde{\nabla}\mathcal{L}_c\|_W^2 = \left(\mathbf{F}_c^{-1} \nabla \mathcal{L}_c\right)^\top \mathbf{F}_c \left(\mathbf{F}_c^{-1} \nabla \mathcal{L}_c\right) = (\nabla \mathcal{L}_c)^\top \mathbf{F}_c^{-1} \nabla \mathcal{L}_c.
\end{equation}
We thus define the \textbf{Riemannian curvature energy} as:  
\begin{equation}
    I_c = \frac{1}{2} (\nabla \mathcal{L}_c)^\top \mathbf{F}_c^{-1} \nabla \mathcal{L}_c.
\end{equation}
Physical and Geometric Interpretation  
The formula can be interpreted through two key perspectives:  
(1) Balance between Gradient and Metric Scaling:  
The numerator \( \nabla \mathcal{L}_c \) represents the gradient magnitude of channel \( c \), indicating the local sensitivity of the loss function to parameter changes. The matrix \( \mathbf{F}_c^{-1} \), the inverse of the Fisher information matrix, characterizes the \emph{local metric scaling} of the parameter space. When the eigenvalues of \( \mathbf{F}_c \) are large in a particular direction (signifying high local curvature or strong parameter correlations), the corresponding components of \( \mathbf{F}_c^{-1} \) are small, which suppresses the gradient contribution in that direction. This mechanism avoids misjudgments of importance caused by geometric distortion in the parameter space. In contrast, the Euclidean norm implicitly assumes \( \mathbf{F}_c = \mathbf{I} \) (the identity matrix), ignoring the actual geometric structure of the manifold.  

(2) Impact of Unit Geometric Perturbations:  
The Riemannian norm of the negative natural gradient measures the effect of \emph{unit-length perturbations in the Riemannian manifold} on the loss function. By normalizing with \( \mathbf{F}_c^{-1} \), perturbations in different parameter directions are converted to the manifold's intrinsic geometric scale. In high-curvature directions (where \( \mathbf{F}_c \) has large eigenvalues), the same Euclidean-distance perturbation corresponds to a smaller effective geometric distance in the manifold, leading to a weaker impact on the loss. This property enables \( I_c \) to accurately capture the dynamic importance of channel parameters in the true geometric space, overcoming the scale biases inherent in the Euclidean framework.

\subsection{Bit Allocation Principle}\label{sec:appendix_bit_allocation}
Following the inference from \cite{rate_distortion}, the quantization error can be approximated as uniformly distributed, and the quantization distortion \( D \) has an exponential relationship with the number of quantization bits \( b \), expressed as:
\begin{equation}
    D \propto 2^{-2b}.
\end{equation}
In our framework, for each channel \( W_c \), the quantization distortion is influenced not only by the number of quantization bits but also by the sensitivity of the channel to the loss function. This sensitivity is measured by \( I_c \geq 1 \) (where we ensure non-negativity by shifting \( I_c \leftarrow I_c + 1 \)). Intuitively, \( I_c \) reflects that even with the same magnitude of quantization error, the impact on the final loss can vary: a larger \( I_c \) indicates that the channel has stronger gradients and higher curvature, meaning the quantization error is amplified and causes greater loss degradation; conversely, a smaller \( I_c \) suggests the channel is less sensitive to errors, resulting in smaller loss impact.

Therefore, the quantization distortion for channel \( c \) can be formulated as:
\begin{equation}
    D_c(b_c) \propto I_c \cdot 2^{-2b_c}.
\end{equation}
Under the global bit budget constraint \(\sum_{c=1}^C b_c = B_{\max}\), where \( B_{\max} \) denotes the total bit allocation across all channels (e.g., \( B_{\max} = 2 \times C \) for 2-bit quantization), we aim to minimize the overall distortion. To this end, we construct the following Lagrangian:
\begin{equation}
    \mathcal{L}(\{b_c\}, \gamma) = \sum_{c=1}^C I_c \cdot 2^{-2b_c} + \gamma\left(\sum_{c=1}^C b_c - B_{\max}\right).
\end{equation}
Taking the partial derivative of \( \mathcal{L} \) with respect to each \( b_c \) and setting it to zero (ignoring discretization effects), we obtain:
\begin{equation}
    \frac{\partial \mathcal{L}}{\partial b_c} = -2 \ln 2 \cdot I_c \cdot 2^{-2b_c} + \gamma = 0.
\end{equation}
which leads to:
\begin{equation}
    2^{-2b_c} = \frac{\gamma}{2\ln2 \cdot I_c}.
\end{equation}
Taking the logarithm on both sides, we further derive:
\begin{equation}
    b_c \propto \log_2(I_c).
\end{equation}
Considering the overall bit budget \( B_{\max} \), the final bit allocation formula is:
\begin{equation}
    b_c = B_{\max} \cdot \frac{\log_2 I_c}{\sum_{c=1}^C \log_2 I_c}.
\end{equation}

\subsection{Approximation of FIM}\label{sec:appendix_fim}
Given the extremely high complexity of directly computing the full FIM (scaling with the square of the weight dimension), we adopt Kronecker factorization approximation (instead of diagonalization) to decompose the FIM into the tensor product of two low-dimensional matrices: $F \approx F_O \otimes F_I$, where $F_O \in \mathbb{R}^{m \times m}$ (output channel FIM) and $F_I \in \mathbb{R}^{n \times n}$ (input channel FIM). This decomposition is implemented as follows:  
\begin{enumerate}
    \item Gradient definition: $\nabla_{W} \ell$ denotes the gradient of the loss function with respect to the weight $W$, computed for a single sequence $s$.
    \item Calculation of $F_I$: Input channel statistics are estimated via the expectation of the outer product of gradients:  
    $$F_I = \frac{1}{m} \cdot \mathbb{E}_{s \sim \mathcal{D}} \left[ \left( \nabla_W \ell \right)^T \cdot \left( \nabla_W \ell \right) \right].$$
    \item Calculation of $F_O$: Output channel statistics are estimated via the expectation of the outer product of gradients:  
    $$F_O = \frac{1}{n} \cdot \mathbb{E}_{s \sim \mathcal{D}} \left[ \left( \nabla_W \ell \right) \cdot \left( \nabla_W \ell \right)^T \right].$$
\end{enumerate}

\subsection{Additional Experiments on the Projection Hyperparameter $\lambda$}
\label{sec:appendix_lambda}

We conducted further ablation studies to examine the sensitivity of the projection hyperparameter $\lambda$ across different model architectures under the 2-bit quantization setting on the Wikitext2 test set. Specifically, we tested three representative models: LLaMA2-7B, LLaMA3-8B, and Qwen2.5-7B. The results are summarized in Table~\ref{tab:lambda_ablation}. 

Overall, we find that $\lambda$ values in the range $[0.01, 0.1]$ consistently yield strong performance across all tested models. Based on these results, we recommend setting $\lambda = 0.05$ as a robust default choice. 
\begin{table}[h]
\centering
\caption{Ablation study of projection hyperparameter $\lambda$ under 2-bit quantization on Wikitext2. Numbers denote perplexity (lower is better).}
\label{tab:lambda_ablation}
\begin{tabular}{lccccccccc}
\toprule
Model & 0 & 0.001 & 0.01 & 0.05 & 0.1 & 0.2 & 0.4 & 0.6 & 0.8 \\
\midrule
LLaMA2-7B   & 9.20 & 7.51 & 5.94 & 5.81 & 5.84 & 5.87 & 6.03 & 6.90 & 13.10 \\
LLaMA3-8B   & 13.32 & 11.19 & 9.17 & 8.79 & 8.78 & 8.99 & 14.86 & 17.63 & 20.92 \\
Qwen2.5-7B  & 13.17 & 10.37 & 8.75 & 8.77 & 9.02 & 11.27 & 15.90 & 19.71 & 36.74 \\
\bottomrule
\end{tabular}
\end{table}

\subsection{Ablation Studies on Group Size and Codebook Vector Length}
\label{sec:appendix_group_vector}
We provide additional ablation studies to analyze the sensitivity of RSAVQ to two critical implementation parameters: the number of groups in WCSG and the codebook vector length. These results complement the main text and justify our chosen default configurations. 
\paragraph{Effect of group size}
We conducted experiments on the LLaMA2-7B model with WikiText2 (sequence length 4096, 2-bit quantization, vector length 6). Results in Table~\ref{tab:group_size_ablation} show that performance improves as the number of groups increases, but the gain diminishes after 4 groups. Specifically, perplexity (ppl) drops significantly from 6.03 to 5.81 when increasing groups from 2 to 4, while further increases (6–10 groups) yield stable results around 5.78–5.79. This indicates moderate sensitivity to group size, with stable performance at $\geq 4$ groups. We thus adopt 4 groups as the default configuration to balance performance and efficiency. 
\begin{table}[h]
\centering
\caption{Perplexity (ppl) of LLaMA2-7B with different WCSG group sizes under 2-bit quantization.}
\label{tab:group_size_ablation}
\begin{tabular}{lcccccc}
\toprule
Groups & 2 & 3 & 4 & 6 & 8 & 10 \\
\midrule
ppl    & 6.03 & 5.88 & 5.81 & 5.79 & 5.78 & 5.78 \\
\bottomrule
\end{tabular}
\end{table}
\paragraph{Effect of codebook vector length}
Vector length plays a central role in vector quantization. We analyzed its impact using LLaMA2-7B on WikiText2 (sequence length 4096, 2-bit quantization, 2 groups for product quantization, 4 groups for WCSG). Table~\ref{tab:vector_length_ablation} shows that longer vector lengths yield slight performance gains (e.g., ppl decreases from 5.81 at length 6 to 5.62 at length 14). However, increasing vector length also raises storage and quantization costs. For example, at 2-bit quantization, vector length growth from 6 to 14 increases the average bit count from 2.0 to 2.875, with corresponding bandwidth costs. 

This demonstrates that RSAVQ’s sensitivity to vector length lies in the performance–cost trade-off: longer vectors offer better performance but incur higher costs. To balance accuracy and efficiency, we use a vector length of 6 in main experiments. 

\begin{table}[h]
\centering
\caption{Impact of codebook vector length on perplexity (ppl) and average bit count under 2-bit quantization (LLaMA2-7B, WikiText2).}
\label{tab:vector_length_ablation}
\begin{tabular}{lcccccc}
\toprule
Vector length & 4 & 6 & 8 & 10 & 12 & 14 \\
\midrule
Avg. bits     & 2.00 & 2.00 & 2.00 & 2.04 & 2.19 & 2.88 \\
ppl           & 5.97 & 5.81 & 5.81 & 5.81 & 5.75 & 5.62 \\
\bottomrule
\end{tabular}
\end{table}
The above results confirm that RSAVQ is moderately sensitive to group size and vector length. Performance saturates after 4 groups, and vector length offers a tunable trade-off between accuracy and cost. The chosen defaults (group size = 4, vector length = 6) provide a balanced configuration for our main experiments. 

\subsection{Quantization Time Comparison}\label{sec:appendix_quant_time}
We have supplemented data on the efficiency of the offline quantization process, comparing RSAVQ with mainstream PTQ methods (VPTQ, GPTVQ) under a 2-bit configuration on an 80GB A100 GPU. Our method adopts a product quantization scheme, which not only reduces quantization time and codebook size compared to conventional vector quantization, but also achieves quantization latency comparable to GPTVQ, where the $vector\_length=1$.
\begin{table}[h]
\centering
\caption{Quantization time comparison of different methods on LLaMA2.}\label{tab:llama_gpu_hours}
\begin{tabular}{lccc}
\hline
 & LLaMA2-7B & LLaMA2-13B & LLaMA2-70B \\
\hline
VPTQ   & 2 GPU hours   & 3.5 GPU hours & 19 GPU hours \\
GPTVQ  & 1 GPU hours   & 1.8 GPU hours & 8 GPU hours \\
RSAVQ  & 1.2 GPU hours & 2 GPU hours   & 10 GPU hours \\
\hline
\end{tabular}
\end{table}

\clearpage
\subsection{Algorithm}\label{sec:appendix_algorithm1}
\algrenewcommand{\algorithmicrequire}{\textbf{Input:}}
\algrenewcommand{\algorithmicensure}{\textbf{Output:}}

\newcommand{\Module}[2]{%
  \vspace{0.8em}%
  \Statex \colorbox{#1!10}{\parbox{\linewidth-2\fboxsep}{\textbf{#2}}}%
}

\begin{algorithm}[H]
\caption{RSAVQ: Riemannian Sensitivity-Aware Vector Quantization}
\label{alg:rsavq_final}
\begin{algorithmic}[1]
\Require{Original weight matrix \( W \in \mathbb{R}^{M \times N} \), total bit budget \( B_{\max} \), sub-vector length \( v \), iterations \( T \), small positive constants \( \lambda \), number of groups \( G \), Fisher matrices \( \{\mathbf{F}_c\}_{c=1}^C \)}
\Ensure{Quantized codebook \( \mathcal{C} \), index matrix \( \mathcal{I} \), original channel positions \( \mathcal{P} \)}

\Module{blue} {\textbf{CWSG: Channel-wise Sensitivity Guidance}}
\State Initialize: \( \mathcal{P}[c] = c \) (preserve original channel order)
\State Compute channel importance scores using Riemannian metric:
\For{ \( c = 1 \) to \( C \)}
    \State Compute natural gradient: \( \tilde{\nabla}\mathcal{L}_c = \mathbf{F}_c^{-1} \nabla\mathcal{L}_c \) \Comment{Eq~\ref{eq:natural_gradient} in paper}
    \State Importance score: \( I_c = \frac{1}{2} \tilde{\nabla}\mathcal{L}_c^\top \mathbf{F}_c \tilde{\nabla}\mathcal{L}_c \) \Comment{Fisher-weighted sensitivity, Eq~\ref{eq:Ic_cal} in paper}
    \State Store \( I_c \) in \( \mathcal{I}_{importance}[c] \)
\EndFor

\State Sort channels by \( \mathcal{I}_{importance} \) (descending) and reorder \( W \), \( \mathcal{P} \)
\State Allocate bits proportionally to sensitivity:
\State \( I[c] = I[c] +1 \)
\For{ \( c = 1 \) to \( C \)}
    \State \( b[c] = \text{Round}\left( B_{\max}\cdot\frac{\log_2I[c]}{\sum_{c = 1}^{C}\log_2I[c]} \right) \)\Comment{Eq~\ref{eq:bit_cal} in paper}
\EndFor

\State Channel grouping: Divide sorted channels into $G$ groups with equal size
    \State $n = \lceil C/G \rceil$
    \For{$g = 1$ to $G$}
        \State Group $g$ channels: $G_g = [(g-1)n+1, \min(gn, C)]$
        \State Average bits per group: $b_g = \text{Round}\left(\frac{\sum_{c\in G_g} b[c]}{|G_g|}\right)$ \Comment{Eq~\ref{eq:bc_group_cal} in paper}
    \EndFor

\Module{red} {\textbf{EDSG: Error Direction Sensitivity Guidance}}
\For{ \( g = 1 \) to \( G \)}
    \State Extract group weights and reshape into sub-vectors: $W_g \in \mathbb{R}^{M\times|G_g|} \rightarrow \{v_{g,l} \in \mathbb{R}^v\}$
    
    \State Initialize codebook \( \mathcal{C}_g \) via K-means with \(2^{b_g[g]}\) cluster centers
    \For{\( t = 1 \) to \( T \)}
        \For{each sub-vector \( v_{g,l} \)}    
            \State Quantize: \( i_{g,l} = \arg\min_i \|v_{g,l} - \mathcal{C}_{g,i}\|^2 \)
            \State Compute error: \( E_{g,l} = v_{g,l} - \mathcal{C}_{g,i_{g,l}} \)
            \State Project error to low-sensitivity direction:
            \State \(  \mathcal{L}_\text{project} \gets \|E_{g, l} + \lambda \cdot \tilde{\nabla}\mathcal{L} \|_{\mathbf{F}}^2  \) \Comment{Eq~\ref{eq:align_loss} in paper}
            \State Update \( \mathcal{C}_g \) via gradient descent on \( \mathcal{L}_{\text{project}} \)
        \EndFor
        \State Store index \( \mathcal{I}[g,l] = i_{g,l} \)
    \EndFor
\EndFor

\State Reconstruct \( \hat{W} \) from \( \mathcal{C} \) and \( \mathcal{I} \)
\State \Return \( \mathcal{C}, \mathcal{I}, \mathcal{P} \)
\end{algorithmic}
\end{algorithm}

\clearpage
\subsection{Additional Experiments}\label{sec:appendix_experiments}
In this section we report additional experimental results for LLaMA-2 models\citep{llama2} and LLaMA-3 models\citep{llama3}.
\begin{table}[htbp]
\centering
\small
\caption{Perplexity on Wikitext-2(Sequence Length=4096) and Zero-Shot Task Accuracy of Various Quantization Algorithms on LLaMA-2 7B.}
\label{tab:llama2-7b}
\resizebox{0.8\textwidth}{!}{
\begin{tabular}{l|l|lllllll}
\hline
\begin{tabular}[c]{@{}l@{}}LlaMA-2 7B \\ seqlen=4096\end{tabular} & bit  & W2↓                                                 & AC & AE   & HE & QA   & WI   & Avg↑  \\ \hline
\rowcolor[HTML]{DEE0E3} 
FP16        & 16    & 5.12  & 43.3  & 76.3  & 57.1  & 78.1  & 68.7  & 64.7  \\ \hline
GPTQ        & 2 & 50.75 & 20.9  & 34.9  & 30.5  & 57.2  & 52.3  & 39.16 \\
GPTVQ       & 2.25  & 6.71  & 31.2  & 66.3  & 46.4  & 72.4  & 64.4  & 56.14 \\
DB-LLM      & 2.01  & 7.23  & 33.53 & 45.2  & 61.98 & 73.18 & 61.72 & 55.12 \\
AQLM        & 2.29  & 6.29  & 34.9  & 66.5  & 50.88 & 74.92 & 65.67 & 58.57 \\
VPTQ        & 2.02  & 6.13  & 35.24 & 63.8  & 52.08 & 75.19 & 64.33 & 58.13 \\
QuIP\#      & 2     & 6.19  & 34.6  & 64.6  & 51.91 & 75.1  & 64.9  & 58.22 \\
\rowcolor[HTML]{DEE0E3} 
RSAVQ(ours) & 2     & 5.81  & 37.2  & 64.4  & 50.71 & 75.4  & 65.74 & 58.6  \\ \hline
GPTQ        & 3     & 8.06  & 31.1  & 58.5  & 45.2  & 71.5  & 59.2  & 53.1  \\
GPTVQ       & 3.125 & 5.44  & 39.93 & 74.07 & 54.21 & 76.17 & 69.06 & 62.69 \\
AQLM        & 3.04  & 5.46  & 38.4  & 68.06 & 54.12 & 76.88 & 66.93 & 60.88 \\
VPTQ        & 3.02  & 5.43  & 39.3  & 69.1  & 54.9  & 77.3  & 68    & 61.72 \\
QuIP\#      & 3     & 5.41  & 39.2  & 68.4  & --    & 77.3  & 66.5  & --    \\
\rowcolor[HTML]{DEE0E3} 
RSAVQ(ours) & 3.01  & 5.26  & 41    & 73    & 54.7  & 76.7  & 68.2  & 62.7  \\ \hline
GPTQ        & 4     & 5.49  & 36.8  & 66.2  & 55.4  & 76.6  & 68.2  & 60.64 \\
GPTVQ       & 4.125 & 5.27  & 42.83 & 75.17 & 56.41 & 77.37 & 69.61 & 62.28 \\
AQLM                                                             & 4.04 & \cellcolor[HTML]{FFFFFF}{\color[HTML]{1F2329} 5.21} & 41 & 70.2 & 56 & 78.2 & 67.3 & 62.54 \\
VPTQ        & 4.01  & 5.26  & 39.7  & 69    & 56    & 78.1  & 67.1  & 61.98 \\
QuIP\#      & 4     & 5.19  & 40.5  & 69.1  & --    & 78.4  & 67.6  & --    \\
\rowcolor[HTML]{DEE0E3} 
RSAVQ(ours) & 4.01   & 5.22  & 42    & 74.7  & 56.3  & 76.9  & 68.2  & 63.62 \\ \hline
\end{tabular}
}
\end{table}
\begin{table}[htbp]
\centering
\small
\caption{Perplexity on Wikitext-2(Sequence Length=4096) and Zero-Shot Task Accuracy of Various Quantization Algorithms on LLaMA-2 13B.}
\label{tab:llama2-13b}
\resizebox{0.8\textwidth}{!}{
\begin{tabular}{l|l|lllllll}
\hline
\begin{tabular}[c]{@{}l@{}}LLaMA-2 13B \\ seqlen=4096\end{tabular} & bit & W2↓ & AC & AE & HE & QA & WI & Avg↑ \\ \hline
\rowcolor[HTML]{DEE0E3} 
FP16 & 16 & 4.57 & 48.2 & 79.5 & 60.1 & 79.1 & 72.2 & 67.82 \\ \hline
GPTQ & 2 & 43.84 & 23.3 & 43.3 & 36 & 61.3 & 54.7 & 43.72 \\
GPTVQ & 2.25 & 5.72 & 38.7 & 73.6 & 51.6 & 75.4 & 68.5 & 61.56 \\
DB-LLM & 2.01 & 6.19 & 38.14 & 51.64 & 68.04 & 75.14 & 64.09 & 59.41 \\
AQLM & 2.18 & 5.41 & 39.42 & 69.15 & 54.68 & 76.22 & 68.43 & 61.58 \\
VPTQ & 2.02 & 5.32 & 40.02 & 71.55 & 56.18 & 77.26 & 66.85 & 62.37 \\
QuIP\# & 2 & 5.35 & 39.5 & 69.3 & 56.01 & 77.3 & 67.7 & 61.96 \\
\rowcolor[HTML]{DEE0E3} 
RSAVQ(ours) & 2 & 5.29 & 41.4 & 72.5 & 56.3 & 75.1 & 68.9 & 62.84 \\ \hline
GPTQ & 3 & 5.85 & 38.48 & 65.66 & 53.47 & 76.5 & 63.93 & 59.61 \\
GPTVQ & 3.125 & 4.8 & 44.45 & 77.23 & 58.18 & 77.8 & 71.98 & 59.63 \\
AQLM & 3.03 & 4.82 & \cellcolor[HTML]{FFFFFF}{\color[HTML]{1F2329} 42.58} & 70.88 & 58.3 & 77.26 & 68.43 & 63.49 \\
VPTQ & 3.03 & 4.79 & 42.32 & 73.99 & 58.42 & 77.64 & 68.67 & 64.21 \\
QuIP\# & 3 & 4.78 & 44 & 72.5 & -- & 78.4 & 69.1 & -- \\
\rowcolor[HTML]{DEE0E3} 
RSAVQ(ours) & 3.01 & 4.74 & 44.9 & 77.7 & 57.5 & 78.1 & 72.4 & 66.12 \\ \hline
GPTQ & 4 & 4.78 & 42.49 & 70.45 & 58.67 & 77.75 & 70.01 & 63.87 \\
GPTVQ & 4.125 & 5.27 & 42.83 & 75.17 & 56.41 & 77.37 & 69.61 & 64.28 \\
AQLM & 3.94 & \cellcolor[HTML]{FFFFFF}{\color[HTML]{1F2329} 4.65} & 44.8 & 73.32 & 59.27 & 78.35 & 69.85 & 65.12 \\
VPTQ & 4.02 & 4.64 & 44.37 & 73.19 & 59.37 & 77.75 & 69.77 & 64.89 \\
QuIP\# & 4 & 4.63 & 45.5 & 73.9 & -- & 78.9 & 69.9 & -- \\
\rowcolor[HTML]{DEE0E3} 
RSAVQ(ours) & 4.01 & 4.72 & 46.5 & 78.8 & 59 & 78.3 & 71.5 & 66.82 \\ \hline
\end{tabular}
}
\end{table}
\begin{table}[htbp]
\centering
\small
\caption{Perplexity on Wikitext-2(Sequence Length=4096) and Zero-Shot Task Accuracy of Various Quantization Algorithms on LLaMA-2 70B.}
\label{tab:llama2-70b}
\resizebox{0.8\textwidth}{!}{
\begin{tabular}{l|l|lllllll}
\hline
\begin{tabular}[c]{@{}l@{}}LLaMA-2 70B\\ seqlen=4096\end{tabular} & bit & W2↓ & AC & AE & HE & QA & WI & Avg↑ \\ \hline
\rowcolor[HTML]{DEE0E3} 
FP16 & 16 & 3.12 & 51.11 & 77.74 & 63.97 & 81.12 & 77.11 & 70.21 \\ \hline
GPTQ & 2 & NaN & 35.8 & 67 & 51.8 & 74.6 & 66.7 & 59.18 \\
GPTVQ & 2.25 & 4.25 & 49.4 & 80.47 & 58.26 & 79.4 & 75.2 & 68.55 \\
DB-LLM & 2.01 & 4.64 & 44.45 & 55.93 & 76.16 & 79.27 & 73.32 & 65.83 \\
AQLM & 2.07 & 3.94 & 47.93 & 77.68 & 61.79 & 80.43 & 75.93 & 68.75 \\
VPTQ & 2.07 & 3.93 & 47.7 & 77.1 & 62.98 & 80.3 & 74.98 & 68.61 \\
QuIP\# & 2 & 3.91 & 48.7 & 77.3 & 62.49 & 80.3 & 75.9 & 68.94 \\
\rowcolor[HTML]{DEE0E3} 
RSAVQ(ours) & 2 & 3.55 & 48.03 & 77.56 & 62.43 & {\color[HTML]{1F2329} 80.19} & 77.03 & 69.05 \\ \hline
GPTQ & 3 & 4.4 & 44.11 & 72.73 & 60 & 78.4 & 71.82 & 65.41 \\
AQLM & 3.01 & 3.36 & \cellcolor[HTML]{FFFFFF}{\color[HTML]{1F2329} 50} & {\color[HTML]{1F2329} 77.61} & 63.23 & 81.28 & 77.19 & 69.86 \\
VPTQ & 3.01 & 3.34 & \cellcolor[HTML]{FFFFFF}{\color[HTML]{1F2329} 48.89} & 77.06 & 63.52 & 80.9 & 77.51 & 69.58 \\
QuIP\# & 3 & 3.35 & 50.9 & 77.7 & -- & 81.4 & 76.4 & -- \\
\rowcolor[HTML]{DEE0E3} 
RSAVQ(ours) & 3 & 3.25 & 50.8 & 78.5 & 63.7 & 81.6 & 77.5 & 70.42 \\ \hline
GPTQ & 4 & 3.35 & 49.15 & 76.81 & 63.47 & 81.23 & 75.61 & 69.25 \\ 
AQLM & 4.14 & 3.19 & 50.68 & 77.31 & 63.69 & 81.5 & 76.48 & 69.93 \\
VPTQ & 4.01 & 3.19 & 49.57 & 78.16 & 63.71 & 81.18 & 76.4 & 69.8 \\
QuIP\# & 4 & \cellcolor[HTML]{FFFFFF}{\color[HTML]{1F2329} 3.18} & 50.6 & 78.1 & -- & 81.4 & 77.1 & -- \\
\rowcolor[HTML]{DEE0E3} 
RSAVQ(ours) & 4.01 & 3.11 & 50.9 & 78.7 & 64 & 81.2 & 76.9 & 70.34 \\ \hline
\end{tabular}
}
\end{table}
\begin{table}[htbp]
\centering
\small
\caption{Perplexity on Wikitext-2(Sequence Length=2048) and Zero-Shot Task Accuracy of Various Quantization Algorithms on LLaMA-3 8B.}
\label{tab:llama3-8b}
\resizebox{0.8\textwidth}{!}{
\begin{tabular}{l|l|lllllll}
\hline
\rowcolor[HTML]{FFFFFF} 
{\color[HTML]{000000} \begin{tabular}[c]{@{}c@{}}LLaMA-3 8B  seqlen=2048\end{tabular}} &
  {\color[HTML]{000000} bit} &
  {\color[HTML]{000000} W2↓} &
  {\color[HTML]{000000} AC} &
  {\color[HTML]{000000} AE} &
  {\color[HTML]{000000} HE} &
  {\color[HTML]{000000} QA} &
  {\color[HTML]{000000} WI} &
  \multicolumn{1}{l}{\cellcolor[HTML]{FFFFFF}{\color[HTML]{000000} Avg↑}} \\ \hline
\rowcolor[HTML]{DEE0E3} 
{\color[HTML]{000000} FP16} &
  {\color[HTML]{000000} 16} &
  {\color[HTML]{000000} 6.14} &
  {\color[HTML]{000000} 50.3} &
  {\color[HTML]{000000} 80.1} &
  {\color[HTML]{000000} 60.2} &
  {\color[HTML]{000000} 79.6} &
  {\color[HTML]{000000} 73.1} &
  {\color[HTML]{000000} 68.66} \\ \hline
\rowcolor[HTML]{FFFFFF} 
{\color[HTML]{000000} GPTQ} &
  {\color[HTML]{000000} 2} &
  {\color[HTML]{000000} 210} &
  {\color[HTML]{000000} 19.9} &
  {\color[HTML]{000000} 28.8} &
  {\color[HTML]{000000} 27.7} &
  {\color[HTML]{000000} 53.9} &
  {\color[HTML]{000000} 50.5} &
  {\color[HTML]{000000} 36.16} \\
\rowcolor[HTML]{FFFFFF} 
{\color[HTML]{000000} DB-LLM} &
  {\color[HTML]{000000} 2} &
  {\color[HTML]{000000} 13.6} &
  {\color[HTML]{000000} 28.2} &
  {\color[HTML]{000000} 59.1} &
  {\color[HTML]{000000} 42.1} &
  {\color[HTML]{000000} 68.9} &
  {\color[HTML]{000000} 60.4} &
  {\color[HTML]{000000} 51.74} \\
\rowcolor[HTML]{FFFFFF} 
{\color[HTML]{000000} QuIP} &
  {\color[HTML]{000000} 2} &
  {\color[HTML]{000000} 85.1} &
  {\color[HTML]{000000} 21.3} &
  {\color[HTML]{000000} 29} &
  {\color[HTML]{000000} 29.2} &
  {\color[HTML]{000000} 52.9} &
  {\color[HTML]{000000} 51.7} &
  {\color[HTML]{000000} 36.81} \\
\rowcolor[HTML]{FFFFFF} 
{\color[HTML]{000000} QuIP\#} &
  {\color[HTML]{000000} 2} &
  {\color[HTML]{000000} 9.11} &
  {\color[HTML]{000000} 39.2} &
  {\color[HTML]{000000} 72.9} &
  {\color[HTML]{000000} --} &
  {\color[HTML]{000000} 75.6} &
  {\color[HTML]{000000} 68.2} &
  {\color[HTML]{000000} --} \\
\rowcolor[HTML]{FFFFFF} 
{\color[HTML]{000000} VPTQ} &
  {\color[HTML]{000000} 2.08} &
  {\color[HTML]{000000} 9.29} &
  {\color[HTML]{000000} 36.9} &
  {\color[HTML]{000000} 71} &
  {\color[HTML]{000000} 52.2} &
  {\color[HTML]{000000} 75.1} &
  {\color[HTML]{000000} 65.9} &
  {\color[HTML]{000000} 60.22} \\
\rowcolor[HTML]{DEE0E3} 
\multicolumn{1}{l|}{\cellcolor[HTML]{DEE0E3}{\color[HTML]{000000} RSAVQ(ours)}} &
  {\color[HTML]{000000} 2} &
  {\color[HTML]{000000} 8.79} &
  {\color[HTML]{000000} 40.9} &
  {\color[HTML]{000000} 74} &
  {\color[HTML]{000000} 51.9} &
  {\color[HTML]{000000} 75.7} &
  {\color[HTML]{000000} 66.1} &
  {\color[HTML]{000000} 61.72} \\ \hline
\rowcolor[HTML]{FFFFFF} 
{\color[HTML]{000000} GPTQ} &
  {\color[HTML]{000000} 3} &
  {\color[HTML]{000000} 8.2} &
  {\color[HTML]{000000} 37.7} &
  {\color[HTML]{000000} 70.5} &
  {\color[HTML]{000000} 54.3} &
  {\color[HTML]{000000} 74.9} &
  {\color[HTML]{000000} 71.1} &
  {\color[HTML]{000000} 61.7} \\
\rowcolor[HTML]{FFFFFF} 
{\color[HTML]{000000} QuIP} &
  {\color[HTML]{000000} 3} &
  {\color[HTML]{000000} 7.5} &
  {\color[HTML]{000000} 41} &
  {\color[HTML]{000000} 72.9} &
  {\color[HTML]{000000} 55.4} &
  {\color[HTML]{000000} 76.8} &
  {\color[HTML]{000000} 72.5} &
  {\color[HTML]{000000} 63.72} \\
\rowcolor[HTML]{FFFFFF} 
{\color[HTML]{000000} QuIP\#} &
  {\color[HTML]{000000} 3} &
  {\color[HTML]{000000} 6.77} &
  {\color[HTML]{000000} 46.4} &
  {\color[HTML]{000000} 77.4} &
  {\color[HTML]{000000} --} &
  {\color[HTML]{000000} 77.9} &
  {\color[HTML]{000000} 72.9} &
  {\color[HTML]{000000} --} \\
\rowcolor[HTML]{FFFFFF} 
{\color[HTML]{000000} VPTQ} &
  {\color[HTML]{000000} 3.03} &
  {\color[HTML]{000000} 6.97} &
  {\color[HTML]{000000} 45.8} &
  {\color[HTML]{000000} 77.5} &
  {\color[HTML]{000000} 58.4} &
  {\color[HTML]{000000} 78.2} &
  {\color[HTML]{000000} 73.4} &
  {\color[HTML]{000000} 66.66} \\
\rowcolor[HTML]{DEE0E3} 
\multicolumn{1}{l|}{\cellcolor[HTML]{DEE0E3}{\color[HTML]{000000} RSAVQ(ours)}} &
  {\color[HTML]{000000} 3.01} &
  {\color[HTML]{000000} 6.34} &
  {\color[HTML]{000000} 46} &
  {\color[HTML]{000000} 76.1} &
  {\color[HTML]{000000} 57.7} &
  {\color[HTML]{000000} 78.2} &
  {\color[HTML]{000000} 73.9} &
  {\color[HTML]{000000} 66.38} \\ \hline
\rowcolor[HTML]{FFFFFF} 
{\color[HTML]{000000} GPTQ} &
  {\color[HTML]{000000} 4} &
  {\color[HTML]{000000} 6.5} &
  {\color[HTML]{000000} 47.7} &
  {\color[HTML]{000000} 78.8} &
  {\color[HTML]{000000} 59} &
  {\color[HTML]{000000} 78.4} &
  {\color[HTML]{000000} 72.6} &
  {\color[HTML]{000000} 67.3} \\
\rowcolor[HTML]{FFFFFF} 
{\color[HTML]{000000} QuIP} &
  {\color[HTML]{000000} 4} &
  {\color[HTML]{000000} 6.5} &
  {\color[HTML]{000000} 47.4} &
  {\color[HTML]{000000} 78.2} &
  {\color[HTML]{000000} 58.6} &
  {\color[HTML]{000000} 78.2} &
  {\color[HTML]{000000} 73.2} &
  {\color[HTML]{000000} 67.12} \\
\rowcolor[HTML]{FFFFFF} 
{\color[HTML]{000000} QuIP\#} &
  {\color[HTML]{000000} 4} &
  {\color[HTML]{000000} 6.34} &
  {\color[HTML]{000000} 50.2} &
  {\color[HTML]{000000} 80.1} &
  {\color[HTML]{000000} --} &
  {\color[HTML]{000000} 79.7} &
  {\color[HTML]{000000} 72.9} &
  {\color[HTML]{000000} --} \\
\rowcolor[HTML]{FFFFFF} 
{\color[HTML]{000000} VPTQ} &
  {\color[HTML]{000000} 4.03} &
  {\color[HTML]{000000} 6.42} &
  {\color[HTML]{000000} 49.1} &
  {\color[HTML]{000000} 78.8} &
  {\color[HTML]{000000} 59.3} &
  {\color[HTML]{000000} 78.7} &
  {\color[HTML]{000000} 74.8} &
  {\color[HTML]{000000} 68.14} \\
\rowcolor[HTML]{DEE0E3} 
\multicolumn{1}{l|}{{\color[HTML]{000000} RSAVQ(ours)}} &
  {\color[HTML]{000000} 4.01} &
  {\color[HTML]{000000} 6.31} &
  {\color[HTML]{000000} 48.5} &
  {\color[HTML]{000000} 79.7} &
  {\color[HTML]{000000} 59.8} &
  {\color[HTML]{000000} 78.9} &
  {\color[HTML]{000000} 75.2} &
  {\color[HTML]{000000} 68.42} \\ \hline
\end{tabular}
}
\end{table}
\begin{table}[ht]
\centering
\small
\caption{Perplexity on Wikitext-2(Sequence Length=2048) and Zero-Shot Task Accuracy of Various Quantization Algorithms on LLaMA-3 70B.}
\label{tab:llama3-70b}
\resizebox{0.8\textwidth}{!}{
\begin{tabular}{l|l|lllllll}
\hline
\rowcolor[HTML]{FFFFFF} 
{\color[HTML]{000000} LLaMA-3 70B seqlen=2048} &
  {\color[HTML]{000000} bit} &
  {\color[HTML]{000000} W2↓} &
  {\color[HTML]{000000} AC} &
  {\color[HTML]{000000} AE} &
  {\color[HTML]{000000} HE} &
  {\color[HTML]{000000} QA} &
  {\color[HTML]{000000} WI} &
  {\color[HTML]{000000} Avg↑} \\ \hline
\rowcolor[HTML]{DEE0E3} 
{\color[HTML]{000000} FP16} &
  {\color[HTML]{000000} 16} &
  {\color[HTML]{000000} 2.9} &
  {\color[HTML]{000000} 60.1} &
  {\color[HTML]{000000} 87} &
  {\color[HTML]{000000} 66.3} &
  {\color[HTML]{000000} 82.4} &
  {\color[HTML]{000000} 80.8} &
  {\color[HTML]{000000} 75.32} \\ \hline
\rowcolor[HTML]{FFFFFF} 
{\color[HTML]{000000} GPTQ} &
  {\color[HTML]{000000} 2} &
  {\color[HTML]{000000} 11.9} &
  {\color[HTML]{000000} 24.6} &
  {\color[HTML]{000000} 38.9} &
  {\color[HTML]{000000} 41} &
  {\color[HTML]{000000} 62.7} &
  {\color[HTML]{000000} 59.9} &
  {\color[HTML]{000000} 45.42} \\
\rowcolor[HTML]{FFFFFF} 
{\color[HTML]{000000} QuIP} &
  {\color[HTML]{000000} 2} &
  {\color[HTML]{000000} 13} &
  {\color[HTML]{000000} 26.5} &
  {\color[HTML]{000000} 48.9} &
  {\color[HTML]{000000} 40.9} &
  {\color[HTML]{000000} 65.3} &
  {\color[HTML]{000000} 61.7} &
  {\color[HTML]{000000} 48.66} \\
\rowcolor[HTML]{FFFFFF} 
{\color[HTML]{000000} QuIP\#} &
  {\color[HTML]{000000} 2} &
  {\color[HTML]{000000} 5.6} &
  {\color[HTML]{000000} 18.3} &
  {\color[HTML]{000000} 32.2} &
  {\color[HTML]{000000} --} &
  {\color[HTML]{000000} 54.7} &
  {\color[HTML]{000000} 68.9} &
  {\color[HTML]{000000} --} \\
\rowcolor[HTML]{FFFFFF} 
{\color[HTML]{000000} VPTQ} &
  {\color[HTML]{000000} 2.07} &
  {\color[HTML]{000000} 5.66} &
  {\color[HTML]{000000} 54.2} &
  {\color[HTML]{000000} 83.6} &
  {\color[HTML]{000000} 61.8} &
  {\color[HTML]{000000} 80.1} &
  {\color[HTML]{000000} 74} &
  {\color[HTML]{000000} 70.74} \\
\rowcolor[HTML]{DEE0E3} 
{\color[HTML]{000000} RSAVQ(ours)} &
  {\color[HTML]{000000} 2} &
  {\color[HTML]{000000} 5.6} &
  {\color[HTML]{000000} 54.4} &
  {\color[HTML]{000000} 83.1} &
  {\color[HTML]{000000} 61.7} &
  {\color[HTML]{000000} 80.2} &
  {\color[HTML]{000000} 77.1} &
  {\color[HTML]{000000} 71.3} \\ \hline
\rowcolor[HTML]{FFFFFF} 
{\color[HTML]{000000} GPTQ} &
  {\color[HTML]{000000} 3} &
  {\color[HTML]{000000} 5.2} &
  {\color[HTML]{000000} 52.1} &
  {\color[HTML]{000000} 79.6} &
  {\color[HTML]{000000} 63.5} &
  {\color[HTML]{000000} 80.6} &
  {\color[HTML]{000000} 77.1} &
  {\color[HTML]{000000} 70.58} \\
\rowcolor[HTML]{FFFFFF} 
{\color[HTML]{000000} QuIP} &
  {\color[HTML]{000000} 3} &
  {\color[HTML]{000000} 4.7} &
  {\color[HTML]{000000} 54.9} &
  {\color[HTML]{000000} 83.3} &
  {\color[HTML]{000000} 63.9} &
  {\color[HTML]{000000} 82.3} &
  {\color[HTML]{000000} 78.4} &
  {\color[HTML]{000000} 72.56} \\
\rowcolor[HTML]{FFFFFF} 
{\color[HTML]{000000} QuIP\#} &
  {\color[HTML]{000000} 3} &
  {\color[HTML]{000000} 3.8} &
  {\color[HTML]{000000} 31.1} &
  {\color[HTML]{000000} 36.6} &
  {\color[HTML]{000000} --} &
  {\color[HTML]{000000} 58.8} &
  {\color[HTML]{000000} 76.4} &
  {\color[HTML]{000000} --} \\
\rowcolor[HTML]{FFFFFF} 
{\color[HTML]{000000} VPTQ} &
  {\color[HTML]{000000} 3.01} &
  {\color[HTML]{000000} 3.81} &
  {\color[HTML]{000000} 57.3} &
  {\color[HTML]{000000} 84.7} &
  {\color[HTML]{000000} 65.5} &
  {\color[HTML]{000000} 81.7} &
  {\color[HTML]{000000} 79.2} &
  {\color[HTML]{000000} 73.68} \\
\rowcolor[HTML]{DEE0E3} 
{\color[HTML]{000000} RSAVQ(ours)} &
  {\color[HTML]{000000} 3} &
  {\color[HTML]{000000} 3.69} &
  {\color[HTML]{000000} 58.1} &
  {\color[HTML]{000000} 85.2} &
  {\color[HTML]{000000} 67} &
  {\color[HTML]{000000} 81.1} &
  {\color[HTML]{000000} 79.9} &
  {\color[HTML]{000000} 74.26} \\ \hline
\rowcolor[HTML]{FFFFFF} 
{\color[HTML]{000000} GPTQ} &
  {\color[HTML]{000000} 4} &
  {\color[HTML]{000000} 3.3} &
  {\color[HTML]{000000} 58.4} &
  {\color[HTML]{000000} 86.3} &
  {\color[HTML]{000000} 66.1} &
  {\color[HTML]{000000} 82.9} &
  {\color[HTML]{000000} 80.7} &
  {\color[HTML]{000000} 74.88} \\
\rowcolor[HTML]{FFFFFF} 
{\color[HTML]{000000} QuIP} &
  {\color[HTML]{000000} 4} &
  {\color[HTML]{000000} 3.4} &
  {\color[HTML]{000000} 58.7} &
  {\color[HTML]{000000} 86} &
  {\color[HTML]{000000} 65.7} &
  {\color[HTML]{000000} 82.5} &
  {\color[HTML]{000000} 79.7} &
  {\color[HTML]{000000} 74.52} \\
\rowcolor[HTML]{FFFFFF} 
{\color[HTML]{000000} QuIP\#} &
  {\color[HTML]{000000} 4} &
  {\color[HTML]{000000} 3.21} &
  {\color[HTML]{000000} 35} &
  {\color[HTML]{000000} 67.3} &
  {\color[HTML]{000000} --} &
  {\color[HTML]{000000} 71.9} &
  {\color[HTML]{000000} 76.7} &
  {\color[HTML]{000000} --} \\
\rowcolor[HTML]{FFFFFF} 
{\color[HTML]{000000} VPTQ} &
  {\color[HTML]{000000} 4.05} &
  {\color[HTML]{000000} 3.15} &
  {\color[HTML]{000000} 59} &
  {\color[HTML]{000000} 86.1} &
  {\color[HTML]{000000} 66.2} &
  {\color[HTML]{000000} 82.4} &
  {\color[HTML]{000000} 79.8} &
  {\color[HTML]{000000} 74.7} \\
\rowcolor[HTML]{DEE0E3} 
{\color[HTML]{000000} RSAVQ(ours)} &
  {\color[HTML]{000000} 4.01} &
  {\color[HTML]{000000} 3.11} &
  {\color[HTML]{000000} 59.2} &
  {\color[HTML]{000000} 86.4} &
  {\color[HTML]{000000} 66.1} &
  {\color[HTML]{000000} 83.4} &
  {\color[HTML]{000000} 80.4} &
  {\color[HTML]{000000} 75.1} \\ \hline
\end{tabular}
}
\end{table}

\clearpage

\section*{NeurIPS Paper Checklist}

\begin{enumerate}

\item {\bf Claims}
    \item[] Question: Do the main claims made in the abstract and introduction accurately reflect the paper's contributions and scope?
    \item[] Answer: \answerYes{}
    \item[] Justification: We provide both theoretical and empirical results supporting the main claim of the paper.

\item {\bf Limitations}
    \item[] Question: Does the paper discuss the limitations of the work performed by the authors?
    \item[] Answer: \answerYes{}
    \item[] Justification: The Appendix~\ref{appendix:limitation} highlights some of the limitation of this work.
    
\item {\bf Theory Assumptions and Proofs}
    \item[] Question: For each theoretical result, does the paper provide the full set of assumptions and a complete (and correct) proof?
    \item[] Answer: \answerYes{}
    \item[] Justification: All statements are provided with proofs.
    
\item {\bf Experimental Result Reproducibility}
    \item[] Question: Does the paper fully disclose all the information needed to reproduce the main experimental results of the paper to the extent that it affects the main claims and/or conclusions of the paper (regardless of whether the code and data are provided or not)?
    \item[] Answer: \answerYes{}
    \item[] Justification: We hope to provide all hyperparameters and experimental details in the Section~\ref{experiments} and provide code to reproduce the experiments. 
    \item[] Guidelines:
    \begin{itemize}
        \item The answer NA means that the paper does not include experiments.
        \item If the paper includes experiments, a No answer to this question will not be perceived well by the reviewers: Making the paper reproducible is important, regardless of whether the code and data are provided or not.
        \item If the contribution is a dataset and/or model, the authors should describe the steps taken to make their results reproducible or verifiable.
        \item Depending on the contribution, reproducibility can be accomplished in various ways. For example, if the contribution is a novel architecture, describing the architecture fully might suffice, or if the contribution is a specific model and empirical evaluation, it may be necessary to either make it possible for others to replicate the model with the same dataset, or provide access to the model. In general. releasing code and data is often one good way to accomplish this, but reproducibility can also be provided via detailed instructions for how to replicate the results, access to a hosted model (e.g., in the case of a large language model), releasing of a model checkpoint, or other means that are appropriate to the research performed.
        \item While NeurIPS does not require releasing code, the conference does require all submissions to provide some reasonable avenue for reproducibility, which may depend on the nature of the contribution. For example:
        \begin{enumerate}
            \item If the contribution is primarily a new algorithm, the paper should make it clear how to reproduce that algorithm.
            \item If the contribution is primarily a new model architecture, the paper should describe the architecture clearly and fully.
            \item If the contribution is a new model (e.g., a large language model), then there should either be a way to access this model for reproducing the results or a way to reproduce the model (e.g., with an open-source dataset or instructions for how to construct the dataset).
            \item We recognize that reproducibility may be tricky in some cases, in which case authors are welcome to describe the particular way they provide for reproducibility. In the case of closed-source models, it may be that access to the model is limited in some way (e.g., to registered users), but it should be possible for other researchers to have some path to reproducing or verifying the results.
        \end{enumerate}

    \end{itemize}

\item {\bf Open access to data and code}
    \item[] Question: Does the paper provide open access to the data and code, with sufficient instructions to faithfully reproduce the main experimental results, as described in supplemental material?
    \item[] Answer: \answerNo{}
    \item[] Justification: We aim to collect the code as soon as possible in a Git repository.
    
\item {\bf Experimental Setting/Details}
    \item[] Question: Does the paper specify all the training and test details (e.g., data splits, hyperparameters, how they were chosen, type of optimizer, etc.) necessary to understand the results?
    \item[] Answer: \answerYes{} 
    \item[] Justification: We provide all experimental details in the main text, as well as the appendix.
    \item[] Guidelines:
    \begin{enumerate}
        \item The answer NA means that the paper does not include experiments.
        \item The experimental setting should be presented in the core of the paper to a level of detail that is necessary to appreciate the results and make sense of them.
        \item The full details can be provided either with the code, in appendix, or as supplemental material.
    \end{enumerate}

\item {\bf Experiment Statistical Significance}
    \item[] Question: Does the paper report error bars suitably and correctly defined or other appropriate information about the statistical significance of the experiments?
    \item[] Answer:\answerNo{}
    \item[] Justification: Many of the experiments were performed on large scale models or foundation models, rendering the computation of multiple seeds unrealistic.

\item {\bf Experiments Compute Resources}
    \item[] Question: For each experiment, does the paper provide sufficient information on the computer resources (type of compute workers, memory, time of execution) needed to reproduce the experiments?
    \item[] Answer: \answerYes{}
    \item[] Justification: We specified the experimental equipment required for our experiments in the experiment section.

\item {\bf Code Of Ethics}
    \item[] Question: Does the research conducted in the paper conform, in every respect, with the NeurIPS Code of Ethics \url{https://neurips.cc/public/EthicsGuidelines}?
    \item[] Answer: \answerYes{}
    \item[] Justification: The paper should be considered a theory and/or conceptual paper. We discussed implication for efficient LLM deployment in the main text, and can not anticipate that the presented results can not conform in any aspect with the NeurIPS Code of Ethics.

\item {\bf Broader Impacts}
    \item[] Question: Does the paper discuss both potential positive societal impacts and negative societal impacts of the work performed?
    \item[] Answer: \answerNA{}
    \item[] Justification: here is no societal impact of the work performed.
    
\item {\bf Safeguards}
    \item[] Question: Does the paper describe safeguards that have been put in place for responsible release of data or models that have a high risk for misuse (e.g., pretrained language models, image generators, or scraped datasets)?
    \item[] Answer: \answerNA{}
    \item[] Justification: No data and models realese.

\item {\bf Licenses for existing assets}
    \item[] Question: Are the creators or original owners of assets (e.g., code, data, models), used in the paper, properly credited and are the license and terms of use explicitly mentioned and properly respected?
    \item[] Answer: \answerYes{}
    \item[] Justification: We clearly cite all code packages and datasets.

\item {\bf New Assets}
    \item[] Question: Are new assets introduced in the paper well documented and is the documentation provided alongside the assets?
    \item[] Answer: \answerNA{}

\item {\bf Crowdsourcing and Research with Human Subjects}
    \item[] Question: For crowdsourcing experiments and research with human subjects, does the paper include the full text of instructions given to participants and screenshots, if applicable, as well as details about compensation (if any)? 
    \item[] Answer: \answerNA{}
    
\item {\bf Institutional Review Board (IRB) Approvals or Equivalent for Research with Human Subjects}
    \item[] Question: Does the paper describe potential risks incurred by study participants, whether such risks were disclosed to the subjects, and whether Institutional Review Board (IRB) approvals (or an equivalent approval/review based on the requirements of your country or institution) were obtained?
    \item[] Answer: \answerNA{}
\item {\bf Declaration of LLM usage}
    \item[] Question: Does the paper describe the usage of LLMs if it is an important, original, or non-standard component of the core methods in this research? Note that if the LLM is used only for writing, editing, or formatting purposes and does not impact the core methodology, scientific rigorousness, or originality of the research, declaration is not required.
    \item[] Answer: \answerNA{}
    \item[] Justification: The core method development in this research does not involve LLMs as any important, original, or non-standard components

\end{enumerate}
\end{document}